\documentclass{article}

\usepackage[preprint,nonatbib]{neurips_2024}

\usepackage[utf8]{inputenc} 
\usepackage[T1]{fontenc}    
\usepackage{hyperref}       
\usepackage{url}            
\usepackage{booktabs}       
\usepackage{amsfonts}       
\usepackage{nicefrac}       
\usepackage{microtype}      
\usepackage{xcolor}         
\usepackage{graphicx}
\usepackage{amsmath}
\usepackage{wrapfig}
\usepackage{float}
\usepackage{diagbox}
\usepackage{subcaption}
\usepackage{multirow}

\title{Interpreting Key Mechanisms of Factual Recall in Transformer-Based Language Models}

\author{
  Ang Lv$^{1}$\thanks{These authors contributed equally. 
  Ang Lv studied the key mechanisms in the GPT-2 and OPT models and led the writing.
  Yuhan Chen refined the code to support the validation of these mechanisms in the Llama-2 models and created the figures. 
  Kaiyi Zhang developed various tasks to test the detected mechanisms. 
  Yuhan Chen and Kaiyi Zhang provided Ang Lv with some valuable insights.
  }
  \quad Yuhan Chen$^{2}$\footnotemark[1] \quad Kaiyi Zhang$^{1}$\footnotemark[1]  \\
  \textbf{Yulong Wang$^{3}$} \quad \textbf{Lifeng Liu$^{3}$} \quad \textbf{Ji-Rong Wen}$^{1}$ \quad \textbf{Jian Xie}$^{3}$\thanks{Corresponding authors.} \quad \textbf{Rui Yan}$^{1}$\footnotemark[2] \\
  $^{1}$ Gaoling School of Artificial Intelligence, Renmin University of China\\
  $^{2}$ XiaoMi AI Lab\\
  $^{3}$ Baichuan Inc.\\
  \texttt{\{anglv, kaiyizhang, jrwen, ruiyan\}@ruc.edu.cn} \\
  \texttt{\{chenyuhan5\}@xiaomi.com} \\
  \texttt{\{wangyulong, liulifeng, richard\}@baichuan-inc.com} \\
  \\
  \url{https://github.com/trestad/Factual-Recall-Mechanism}
  \vspace{-5mm}
}

\begin{document}

\maketitle

\begin{abstract}
In this paper, we delve into several mechanisms employed by Transformer-based language models (LLMs) for factual recall tasks. 
We outline a pipeline consisting of three major steps:
(1) Given a prompt ``The capital of France is,'' task-specific attention heads extract the topic token, such as ``France,'' from the context and pass it to subsequent MLPs.
(2) As attention heads' outputs are aggregated with equal weight and added to the residual stream, the subsequent MLP acts as an ``activation,'' which either erases or amplifies the information originating from individual heads. As a result, the topic token ``France'' stands out in the residual stream.
(3) A deep MLP takes ``France'' and generates a component that redirects the residual stream towards the direction of the correct answer, i.e., ``Paris.'' 
This procedure is akin to applying an implicit function such as ``get\_capital($X$),'' and the argument $X$ is the topic token information passed by attention heads.
To achieve the above quantitative and qualitative analysis for MLPs, we proposed a novel analytic method aimed at decomposing the outputs of the MLP into components understandable by humans.
Additionally, we observed a universal anti-overconfidence mechanism in the final layer of models, which suppresses correct predictions. 
We mitigate this suppression by leveraging our interpretation to improve factual recall confidence.
The above interpretations are evaluated across diverse tasks spanning various domains of factual knowledge, using various language models from the GPT-2 families, 1.3B OPT, up to 7B Llama-2, and in both zero- and few-shot setups.
\end{abstract}

\section{Introduction}
\label{sec:intro}
Mechanistic interpretability research aims to reverse engineer deep neural networks into human-understandable algorithms or mechanisms~\cite{wang2023interpretability,olsson2022context}.
This endeavor holds significant value for various applications, including providing insights into internal reasoning~\cite{dutta2024think,kruthoff2024carrying,wendler2024llamas}, enriching knowledge~\cite{meng2022locating, hernandez2023inspecting}, improving controllability~\cite{yu2023characterizing,chen2024fortify}, guiding future architecture and training paradigm designs~\cite{elhage2022solu,reddy2024the}, and detecting safety issues~\cite{shevlane2023model}.
Despite the remarkable achievements made by language models~\cite{brown2020language,openaichatgptblog,openai2023gpt4,touvron2023llama1,touvron2023llama2} based on Transformer~\cite{vaswani} in natural language understanding and generation, they remain mainly opaque to human, with many of their internal mechanisms understudied.
In this paper, we delve into the mechanisms in factual recall tasks, which represent one of the cornerstone abilities of contemporary language models~\cite{zhao2023survey,zhang2023sirens}.

Merullo et al. \cite{merullo2023mechanism} delineated the behavior of language models in factual recall tasks into two major phases as layers deepen: (1) \textit{Argument Formation}: When presented with a query such as ``The capital of France is,'' the decoded top-probable tokens from the residual stream gradually yield the name of the queried country, in this case, ``France.'' 
The authors likened this process to the model forming the argument $X$ for an implicit function such as ``get\_capital($X$).''
(2) \textit{Function Application}: As layers deepen, the top-probable tokens decoded from the residual stream transition from the country name to the capital name, such as from ``France'' to ``Paris.'' 
The authors described this transition as the application of the implicit function and assumed the crucial role of MLPs in this phase. 

Although these observations lay a valuable groundwork, several unresolved research questions remain, which are essential for achieving a deep understanding of the factual recall mechanisms in language models.
For instance, how does the model extract the argument from the context and then pass it to the so-called ``function?''
What exactly constitutes ``function application,'' and how does it correlate with MLPs?
Moreover, Merullo et al.~\cite{merullo2023mechanism} primarily focused on a one-shot setting. 
What about the mechanisms of language models in basic zero-shot settings or when provided with more shots?

\begin{figure}[t]
    \centering
    \includegraphics[width=0.98\linewidth]{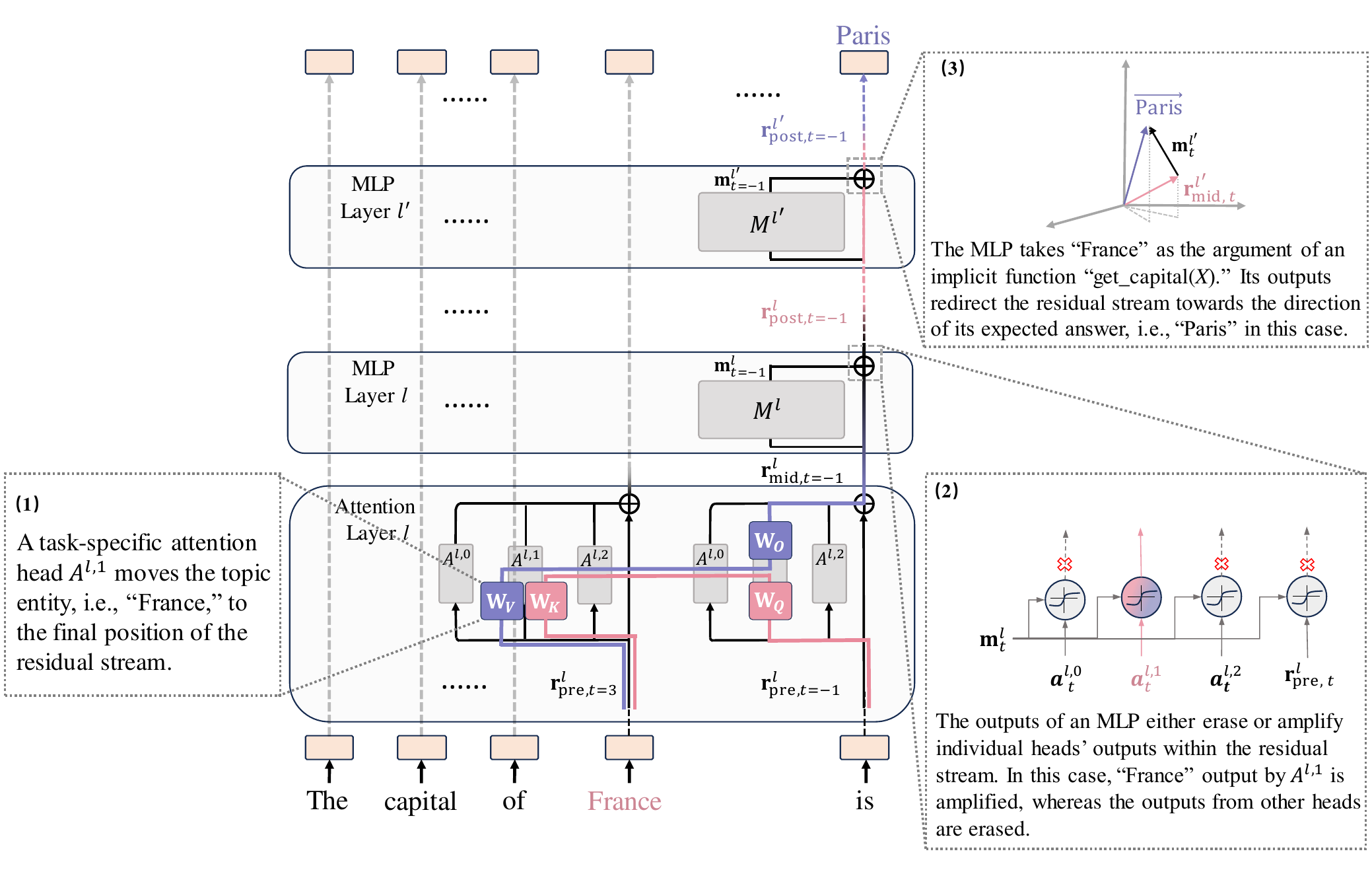}
    \caption{The key mechanisms of factual recall employed by Transformer-based language models. Please refer to \S\ref{sec:background} for detailed notations.}
    \label{fig:circuit_intro}
    \vspace{-5mm}
\end{figure}

In this paper, we investigate the mechanistic interpretability of these crucial questions and offer several innovative insights.
For readers unfamiliar with mechanistic interpretability research, we suggest referring to Section~\ref{sec:background} for necessary backgrounds.
We investigated Transformer-based language models scaling from GPT-2 small, medium, and large~\cite{radford2019language}, 1.3B OPT~\cite{zhang2022opt}, to 7B Llama-2~\cite{touvron2023llama2}, across various factual recall tasks encompassing diverse domains of knowledge. 
Based on our analysis, we outline the following steps by which a language model achieves factual recall.

(1) Some task-specific attention heads in the mid-to-deep layers possess weights inherently sensitive to topic tokens, such as the name of a country, which move these tokens to the final position of the residual stream. 
This mechanism enables the model to extract the ``argument'' from the context and pass it to the later ``function application.'' 
Additionally, some attention heads directly map the argument to the desired outputs, which can be viewed as a partial function application. 

(2) The subsequent MLP serves as an ``activation'' to make the output of task-specific heads stand out in the residual stream, as the output of heads that pass arguments is aggregated with other heads with equal weight.
This MLP can erase or amplify the outputs of individual heads by generating vectors that either align with or oppose the direction of heads' outputs.

(3) Some deep MLP's output incorporates a task-aware component that directs the residual stream towards the direction of the target token's unembedding vector.
The component's addition into the residual stream accomplishes the ``function application.''
The above insights regarding MLPs come from our novel analytical approach, which deconstructs the MLP's output into a linear combination of attention heads' outputs as well as an additional component that encodes the MLP's functionality.

We illustrate these three steps in Figure~\ref{fig:circuit_intro}. 
These mechanisms consistently work in zero-, one-, and few-shot scenarios.
Beyond answering these questions, we have observed that language models universally suppress correct predictions in the final layer. 
This is achieved by integrating frequent tokens into the residual stream via attention heads and employing the MLP to steer the residual stream towards an ``average'' token over the training corpus. 
This phenomenon appears to be a universal anti-overconfidence mechanism adopted by various models to evade significant training loss and persists irrespective of the task, model, and the number of in-context demonstrations.

To sum up, this paper has the following contributions:

1. We examine various models, tasks, and ICL setups, offer a detailed and universal interpretation of the ``argument passing'' and ``function application'' mechanisms employed in Transformer-based language models for factual recall tasks.

2. We propose a novel analysis method to understand certain deep MLPs. 
Our approach reveals that some MLPs behave akin to the activation of attention heads while also generating a task-aware component responsible for ``function application.''
The efficacy of this analysis method is supported by ample empirical evidence.

3. We observed a universal anti-overconfidence mechanism in the final layer of models.
Building on our interpretation, we devise strategies to mitigate this anti-overconfidence mechanism, enhancing factual recall confidence.

\section{Background}
\label{sec:background}

\subsection{A residual-centric perspective of the Transformer}


Within a Transformer layer indexed as $l$, comprising $H$ attention heads, we label each attention head as L$l$H$h$, where $h \in [0,1,..., H-1]$. 
Each attention head's specific operation is denoted by $A^{l,h}(\textbf{r})$, where $\textbf{r}$ is the input residual stream.
Similarly, we denote the MLP as MLP$l$, with its operation as $M^{l}(\textbf{r})$.
Token positions in the sequence dimension are denoted by the subscript $t$.
For clarity, we will omit these subscripts or/and superscripts when there's no ambiguity.

There are three pivotal nodes in the residual stream within a Transformer layer: The first node corresponds to the initial input of the layer, denoted as $\textbf{r}^{l}_{\text{pre}}$. 
The subsequent two nodes are:
\begin{equation}
\begin{aligned}
    &\textbf{r}^{l}_{\text{mid}, t} = \textbf{r}^{l}_{\text{pre}, t} + \sum^{H-1}_{h=0} \textbf{a}^{l,h}_{t}, \quad
    \textbf{r}^{l}_{\text{post}} = \textbf{r}^{l}_{\text{pre}, t} + \sum^{H-1}_{h=0} \textbf{a}^{l,h}_{t} + \textbf{m}^{l}_{t}, \\
    &\text{where} \quad \textbf{a}^{l,h}_{t} = A^{l,h}(\textbf{r}^{l}_{\text{pre}, \leq t}) \quad \text{and} \quad \textbf{m}^{l}_{t} = M^{l}(\textbf{r}^{l}_{\text{mid}, t}).
\end{aligned}
\label{eq:resid}
\end{equation}
Figure~\ref{fig:circuit_intro} illustrates the Transformer layer $l$ from this perspective.
Elhage et al. \cite{elhage2021mathematical} reformulated attention heads for a better theoretical understanding of models, which is equivalent to that presented in \cite{vaswani}.
This reformulation can be represented as follows:
\begin{equation}
    \begin{aligned}
        A(\textbf{r}) = \left(\texttt{softmax}(\textbf{r}^{\top}\textbf{W}^{\top}_{Q}\textbf{W}_{K}\textbf{r}) \otimes \textbf{W}_{O}\textbf{W}_{V}\right) \cdot \textbf{r}.
    \end{aligned}
    \label{eq:ov-qk-formulation}
\end{equation}
Let's denote the OV matrix as $\textbf{W}_{OV} = \textbf{W}_O \textbf{W}_V$ and the QK matrix as $\textbf{W}_{QK} = \textbf{W}^{\top}_Q \textbf{W}_K$, for clarity.
The parameters represented by $\textbf{W}_{OV}$ generate outputs corresponding to inputs;
The parameters represented by $\textbf{W}_{QK}$ generate attention patterns, determining which token's information is moved from and to.

\subsection{Discovery of influential modules}
\paragraph{Causal mediation analysis}
Research indicates that only a few modules (e.g., individual heads, an MLP, etc.) work for a particular task~\cite{wang2023interpretability,merullo2024circuit}. Circuit discovery research aims to uncover these activated modules, grounded in causal mediation analysis~\cite{10.5555/2074022.2074073,NEURIPS2020_92650b2e}.

\begin{wrapfigure}{r}{0.55\linewidth}
\vspace{-7mm}
\begin{center}
    \includegraphics[width=\linewidth]{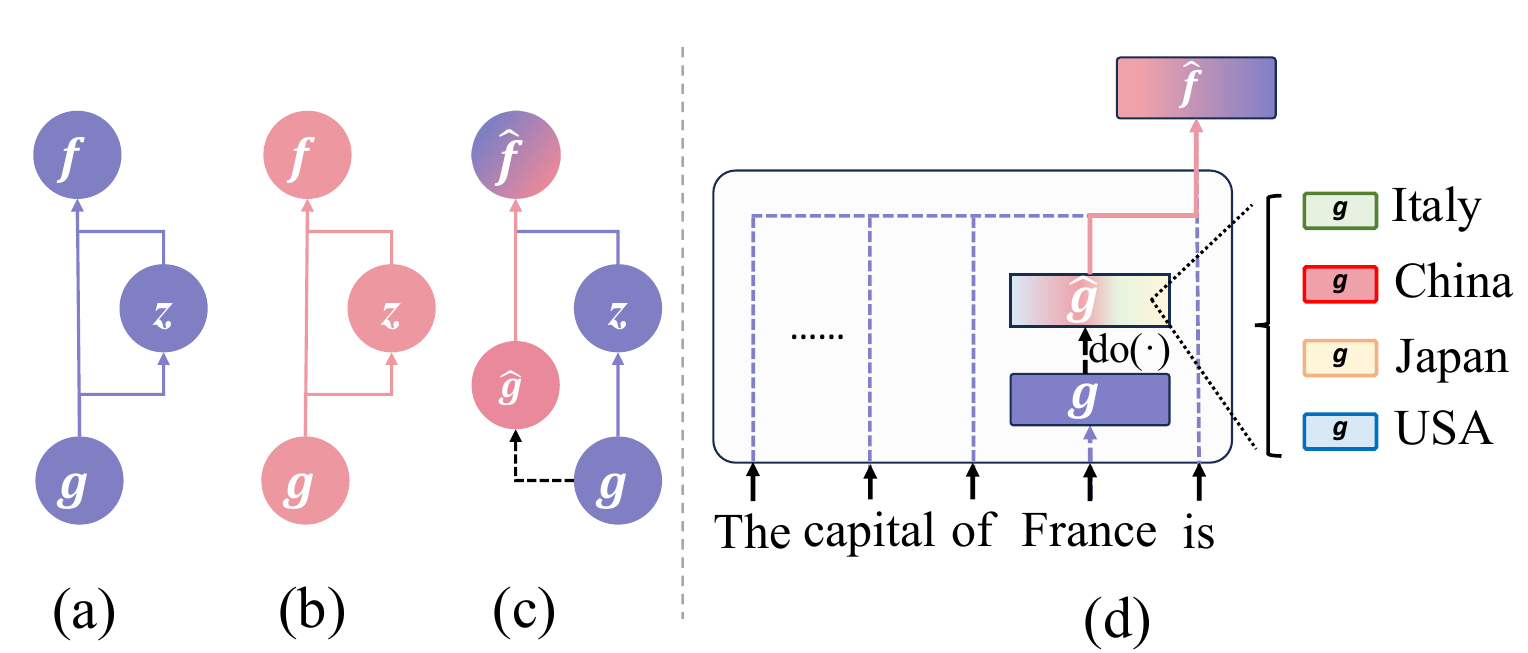}
    \end{center}
    \caption{Subfigures (a), (b), and (c) are toy diagrams of causal mediation analysis in discovering important circuits in a neural network. 
  Colors distinguish different node values.
  Subfigure (d) illustrates an activation patching example in studying the node affecting the correct capital city prediction, as detailed in \S\ref{sec:background2}.}
    \label{fig:causal-mediation}
    \vspace{-2mm}
\end{wrapfigure}

Let us conceptualize the neural network as a causal graph, where each node represents either neurons~\cite{elhage2022solu,elhage2022superposition,makelov2024is} or module activations~\cite{elhage2021mathematical}. 
When there is a direct connection between two nodes in the network, they are linked by an edge in the graph. 
The graph is considered causal because intervening in one node's output can causally influence the subsequent nodes that rely on its output as input. 
A neural network subgraph is termed a circuit~\cite{olah2020zoom}.
In Figure~\ref{fig:causal-mediation}, a toy example elucidates this process.
Suppose we want to discern the impact of node $g$ on node $f$.
When we intervene in $g$'s output (e.g., by knocking out or corrupting it), $f$ is affected by both the change from $g$ and $g$'s indirect intervention via $z$, as shown in Figure~\ref{fig:causal-mediation}(b).
More precisely, we could maintain $z$'s outputs consistent to measure $g$'s ``direct effect'' to $f$, as shown in Figure~\ref{fig:causal-mediation}(c).
By conducting such ablation studies, we can first identify nodes that significant affect the neural network's final outputs.

\paragraph{Activation patching and impact metric}
\label{sec:background2}
``Activation patching'' is a widely employed technique for implementing the interventions in module activations~\cite{wang2023interpretability,meng2022locating,NEURIPS2020_92650b2e}.
Suppose we have an original prompt, denoted as $p_{\text{original}}$, such as ``The capital of $X$ is,'' with the expected answer being $Y$.
The activations of node $f$ in the neural network given this prompt is represented as $f(p_{\text{original}})$. 
Suppose we aim to uncover the node responsible for extracting country identity, which consequently influences the final node $f$, i.e., $\textbf{r}_{\text{post}, t=-1}$ in the final layer. 
A feasible choice is to intervene on $g$, a preceding node of $f$ in the causal graph, at the country name position.

Let us say we have $N$ intervention prompts sharing the template with $p_{\text{original}}$ but differing in keywords. 
For instance, each intervention prompt $p_{\text{intervention},n}$ takes the form ``The capital of $X_n$ is,'' where $X_n$ is sampled from a set of country names distinct from $X$.
By substituting \( g(p_{\text{original}}) \) with intervention information, we corrupt the country identity in \( g(p_{\text{original}}) \). 
This process can be formalized as:
\begin{equation}
    \begin{aligned}
        \hat{g}(p_{\text{original}}) = g\left(p_{\text{original}} \ |\  \texttt{do}(g(p_{\text{original}}) = \frac{1}{N} \sum^{N}_{n=1} g(p_{\text{intervention},n}) )\right).
    \end{aligned}
\end{equation}
The operation $\texttt{do}(\cdot)$, a standard notation in causality~\cite{Pearl_2009}, indicates the replacement of one value with another.
When the node $f$ takes $\hat{g}(p_{\text{original}})$ as inputs, we denote its current value as $\hat{f}$.
If certain metrics indicating the probability of $Y$ decoded from $\hat{f}$ are lower than those from $f$, we can deduce that the edge from $g$ to $f$ positively impacts correct predictions, and vice versa.
The above patching procedure is illustrated in Figure~\ref{fig:causal-mediation}(d).

A feasible metric is the difference in logit values, $\pi(Y, f)$, for our target token $Y$, decoded from node $f$ before and after patching node $g$. 
This metric, denoted $\Delta\pi(Y, f)$, can be formalized as follows:
\begin{equation}
    \begin{aligned}
        \quad \quad \quad \Delta\pi(Y, f)_{t} = (\frac{\pi(Y, \hat{f})_{t}}{\pi(Y, f)_{t}} - 1) \cdot 100\%,
    \end{aligned}
     \label{eq:metric}
\end{equation}
where $\pi (Y, f)_{t}$ is defined as:
\begin{equation}
\begin{aligned}
    \pi (Y, f)_{t} = \left(\texttt{Unembed}_{t}(f)[Y] - \frac{1}{V}\sum^{V}_{v} \texttt{Unembed}_{t}(f)[v]\right),
\end{aligned}
\label{eq:logit}
\end{equation}
and $\texttt{Unembed}(\cdot)$ is computed using $\textbf{W}_{U}\texttt{LN}(\cdot)$.
Here, $\texttt{LN}(\cdot)$ represents the normalization in the final layer. 
$\textbf{W}_{U} \in \mathbb{R}^{V \times d_{\text{model}}}$, where $V$ is the vocabulary size, denotes the unembedding matrix (i.e., the weight of the language model head). 
The notation [$Y$] follows Python syntax, indicating the selection of $Y$-th item.
The subscript $t$ indicates the focus position, set to $-1$ by default in this paper, representing the last position where the current prediction occurs. 
Without ambiguity, this subscript is omitted.
In Eq.~\ref{eq:logit}, we center the logits to measure the net effect of path patching.

\section{How does the language model do factual recall tasks in zero-shot}

Recall that in~\cite{merullo2023mechanism}, authors observed that during factual recall, the language model first forms an ``argument.''
Subsequently, it passes the argument to an implicit function for deriving the final answer.
In this section, we explore the following unresolved research questions within a zero-shot scenario:

1. How does the model discern the ``argument'' from context and pass it to the ``implicit function?''

2. What exactly is the ``function application,'' and how does it relate to MLPs?

\subsection{Experiment setup}

We study Transformer-based language models, scaling from the 117M GPT-2 small~\cite{radford2019language}, GPT-2 medium and large, OPT-1.3B~\cite{zhang2022opt}, to the Llama-2 7B~\cite{touvron2023llama2}.
Our study involved two factual recall tasks concerning country-capital and product-developer associations, requiring different domain knowledge.
The main text details our experiments on GPT-2 small, which has 12 layers and 12 heads in each layer, using the country-capital task. 
Other experiments are presented in appendices.
\textit{Unless specified, the interpretations below are generally validated across models and tasks.}

For the country-capital task, we prompt the language model, asking it to recall the capital of a given country. 
For instance, a prompt would be framed as ``As we all know, the capital of $X$ is.''
Here, $X$ denotes the country in question, and we denote the corresponding capital name of $X$ by $Y$.
To ensure the robustness of our experiments and to mitigate any bias stemming from specific prompt templates, we devise a set of 15 distinct prompts.
We construct a set containing 15 country-capital pairs.
To generate a test dataset, we fill each template with various country names, resulting in a dataset comprising $N=225$ samples.
Data details can be found in Appendix~\ref{apx:dataset}.

\begin{figure}
    \centering
    \includegraphics[width=0.96\linewidth]{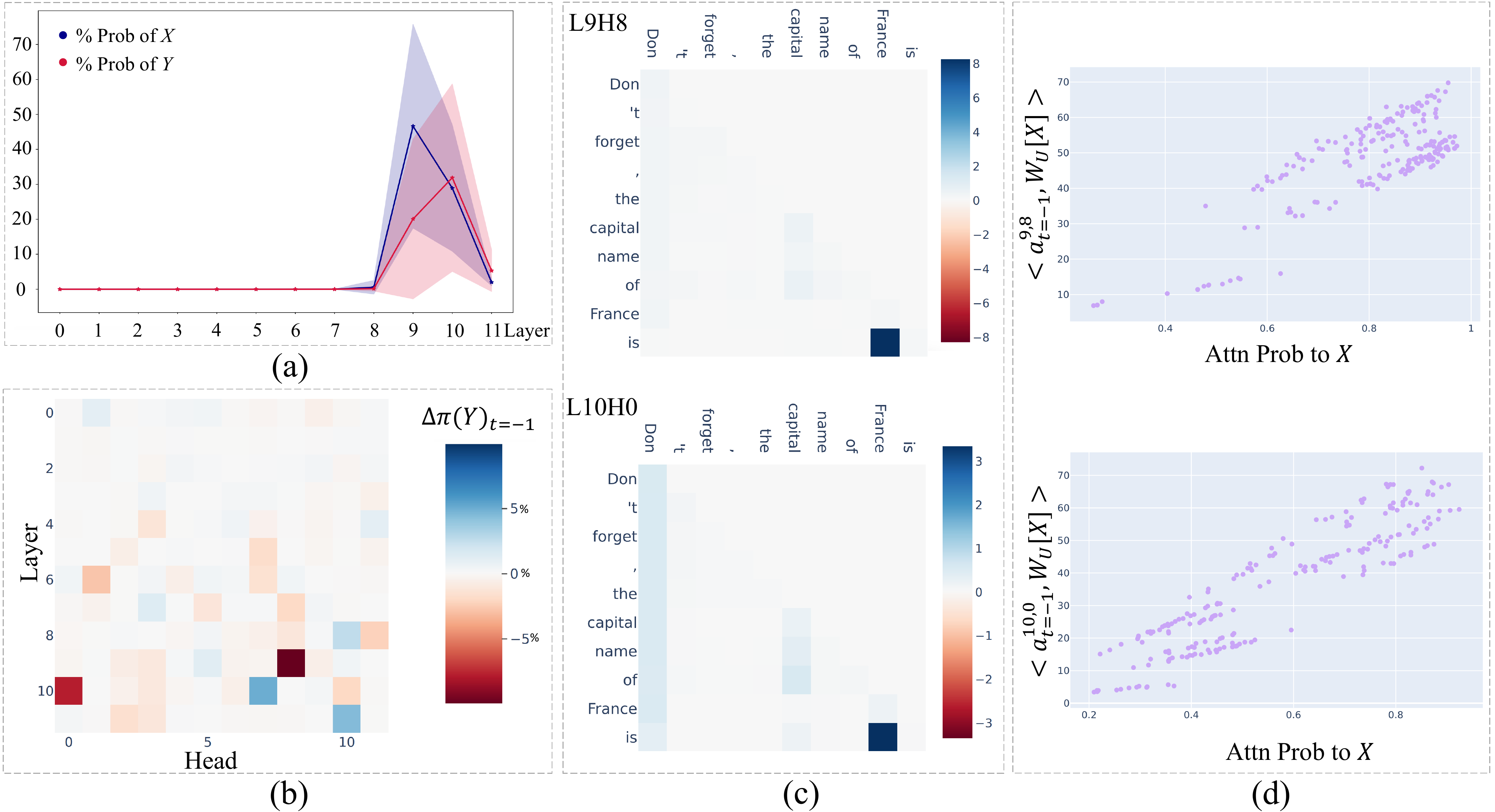}
    \caption{(a) Probabilities of $X$ and $Y$ decoded at each layer. 
    The shaded regions indicate variances. 
    (b) The effect of patching $\textbf{a}^{l,h}_{t=-1} \rightarrow \textbf{r}^{11}_{\text{post},t=-1}$. 
    (c) Value weighted attention pattern of L9H8 and L10H0.
    (d) L9H8 and L10H0's attention to $X$ is proportional to the projection value of their output along $\textbf{W}_{U}[X]$, indicating they are moving country names to the final position. }
    \label{fig:effect-to-logits}
    \vspace{-5mm}
\end{figure}

\subsection{Attention heads pass arguments to the implicit function}
\label{sec:attention-pass-arguments}

\paragraph{L9H8 and L10H0 are argument passers}
We have two closely related pieces of evidence suggesting that attention heads, L9H8 and L10H0, play pivotal roles in the two phases of factual recall.

(1) We replicated experiments conducted by Merullo et al. \cite{merullo2023mechanism} using GPT-2 small. 
For a given input $p_{\text{original}}$, we assessed the early-decoded logits $\pi (X, \textbf{r}^{l}_{\text{post}})$ and $\pi (Y, \textbf{r}^{l}_{\text{post}})$ at each layer $l$, followed by applying softmax to these logits to obtain the probabilities of $X$ and $Y$ at layer $l$ respectively.
Illustrated in Figure \ref{fig:effect-to-logits}(a), at layer 9, $X$ becomes noticeable in the residual stream, marking the onset of ``argument formation.''
At layer 10, the probability of $Y$ surpasses that of $X$ for the first time, indicating the ``function application.'' 

(2) We implement the ``path patching'' algorithm, an activation patching algorithm introduced by \cite{wang2023interpretability} to identify influential nodes.
For patching, \( p_{\text{original}} \) is sampled from the test dataset, while the set of interventions \( \{p_{\text{intervention},n \leq N}\} \) is the entire test dataset.
Taking $\Delta \pi$ (Eq.~\ref{eq:metric}) as the impact metric, path patching results illustrated in Figure \ref{fig:effect-to-logits}(b) indicate that paths from outputs of L9H8 and L10H0, i.e., $\textbf{a}^{9,8}$ and $\textbf{a}^{10,0}$, to $\textbf{r}^{11}_{\text{post}}$ have the most substantial positive impact on correct prediction.

We identified that L9H8 and L10H0 are ``mover heads''~\cite{wang2023interpretability,lieberum2023does,merullo2024circuit}, which move the information from any position they focus on to the final position.
Figure~\ref{fig:effect-to-logits}(c) shows that both L9H8 and L10H0 predominantly attend to the token $X$;
Figure~\ref{fig:effect-to-logits}(d) illustrates the attention score from the final token to $X$, plotted against the inner product $<\textbf{a}^{l,h}_{t=-1}, \textbf{W}_{U}[X]>$. 
The significant correlation between the attention score and the inner product (with Pearson correlation coefficient $0.77$ for L9H8 and $0.90$ for L10H0) strongly suggests that these two heads are responsible for passing the country name, the ``argument'' in the country-capital task, to the final position where the current prediction occurs. 

\paragraph{L9H8 can map country names to capital cities}
While examining L9H8, we decoded $(\textbf{W}^{\text{L9H8}}_{OV}\cdot \textbf{r}^{9}_{\text{pre},t=X})$ by employing $\texttt{Unembed}(\cdot)$, and analyzed 5 tokens with the top logit values.
We noticed that correct capital cities emerged among these top-probable tokens.
This indicates that partial function application happens within this attention head.
We found that mapping countries to capitals is an inherent ability of L9H8: Despite receiving the embedding vector of a country rather than a representation from deeper layers at position $ X $, its OV matrix still yields some capital cities. 
Our investigation further reveals that L9H8 is a task-specific head mainly activated given country-related prompts, as evidenced by its inactivation in other tasks (see Appendix~\ref{apx:more-task}).
In contrast, although L10H0 lacks an inherent ability to map countries to capital cities within its OV matrix, it is activated across a variety of tasks, even those solely reliant on in-context learning~\cite{wang2023interpretability,merullo2024circuit}. 
It functions more like an amplifier for L9H8 or, more generally, an amplifier for the confident answer in the residual stream.

\subsection{MLP acts as the ``activation'' of attention heads}
In \S\ref{sec:attention-pass-arguments}, we have demonstrated that only a few attention heads have a positively direct effect on the correct prediction.
This naturally leads to a question: Given that Transformer models aggregate the outputs of all attention heads with equal weight (as shown in Eq. \ref{eq:resid}), how does the model discern and extract information written by specific heads within the residual stream?

We evaluated the norm of outputs from each attention head and applied principal component analysis (PCA) to identify any anomalous outputs but found no meaningful results.
Then, we hypothesize an ``activation'' mechanism after these attention heads that either amplifies or suppresses the outputs of certain attention heads. 
We turn to the subsequent module following the attention heads—the MLP.

Recall that a Transformer layer's output $\textbf{r}^{l}_{\text{post}}$ is the linear combination of $\textbf{r}^{l}_{\text{pre}}$, the collective output of attention heads $\sum^{H-1}_{h=0} \textbf{a}^{l,h}$, and the MLP output $\textbf{m}^{l}$ (See Eq. \ref{eq:resid}). 
We consider whether $\textbf{m}^{l}$ is equivalent to a linear combination of $\textbf{r}^{l}_{\text{pre}}$ and $\sum^{H-1}_{h=0} \textbf{a}^{l,h}$, where each term's coefficient indicates the degree of amplification or erasure imposed by the MLP.
We assume the linear independence of all $\textbf{a}^{l,h}$ and $\textbf{r}^{l}$, and tackle the following multivariable linear regression problem:
\begin{equation}
    \begin{aligned}
        \Tilde{\textbf{m}}^{l} = \textbf{b}^{l} + \sum^{H-1}_{h=0} w^{l,h} \textbf{a}^{l,h} + w^{l,r} \textbf{r}^{l}_{\text{pre}}.
    \end{aligned}
    \label{eq:lr}
\end{equation}
Here, $w^{l,\cdot}$ represents scalar coefficients to be solved, and the intercept $\textbf{b}^{l}\in \mathbb{R}^{d_{\text{model}}}$ is expected to encapsulate higher-order transformations of head outputs, along with abstract information such as knowledge introduced by the MLP.
We employ the gradient descent method to solve this linear regression problem, with the objective of minimizing the Mean Squared Error (MSE) loss between $\Tilde{\textbf{m}}^{l}$ and $\textbf{m}^{l}$.
We set the learning rate to 0.005 and performed 60,000 optimization steps using the SGD optimizer with a momentum of 0.99.
A 4-fold cross-validation to validate the regression solution. 

The optimization process consistently converges to similar solutions across all layers with minimal loss, regardless of the initialization of $w^{l,h}$, $w^{l,r}$, and $\textbf{b}^{l}$ (refer to Appendix \ref{apx:training} for training specifics). 
This consistency suggests a near-uniqueness of solutions, supporting the assumption of linear independence. 
To further verify the fidelity of the regression solution, additional experiments were conducted. 
We substituted the MLP outputs $\textbf{m}^{l}$ in each layer with $\Tilde{\textbf{m}}^{l}$ calculated using Eq~\ref{eq:lr}. 
Subsequently, we evaluated the Kullback-Leibler (KL) divergence~\cite{kullback1951information} between the output distributions of this modified model and those of the original model, resulting in a minimal KL divergence of 0.21.
Furthermore, the average final logits of $Y$ produced by two models yielded values of 13.08 and 13.15.
These results demonstrate the alignment of the two outputs and thus affirm the fidelity of the regression solution.

We primarily focus on the regression coefficients for layer 9 (detailed in Table~\ref{tab:lr}) as MLP9 immediately follows the L9H8, the most crucial argument passer in the country-capital task.
Notably, among all coefficients, only three exhibit positive values, with $w^{9,8}$ being the highest. 
This suggests that MLP9 enhances the information sourced from L9H8. 
Consequently, the arguments passed by L9H8, along with some capital names it generates, ``stand out'' in the residual stream.
\begin{table}[t]
\small
    \centering
    \renewcommand{\arraystretch}{1.3}
    \caption{Learned coefficients for layer 9.}
    \resizebox{0.98\linewidth}{!}{
    \begin{tabular}{ccccccccccccc}
    \toprule
        $w^{9,0}$ & $w^{9,1}$ & $w^{9,2}$ & $w^{9,3}$ & $w^{9,4}$ & $w^{9,5}$ & $w^{9,6}$ & $w^{9,7}$ & $w^{9,8}$ & $w^{9,9}$ & $w^{9,10}$ & $w^{9,11}$ & $w^{9,r}$\\\hline
         -0.05 & -0.24 & 0.04 & -0.01 & -0.26 & -0.13 & -0.17 & -0.04 & 0.08 & -0.09 & -0.08 & -0.64 & 0.05\\\bottomrule
    \end{tabular}}
    \label{tab:lr}
    \vspace{-3mm}
\end{table}

Thus far, we demonstrated that task-specific heads pass the arguments, and a subsequent MLP serves as an ``activation'' for the attention outputs, effectively highlighting the expected arguments in the residual stream.
These findings collectively address the first research question. 

\subsection{Function application is ``\ \texorpdfstring{$\textbf{b} + \textbf{r}_{\textnormal{mid}}$}{}\ ''}
\label{sec:proj}
While partial function application has been identified within the OV circuit of L9H8, the probability of capital cities $Y$ by the end of layer 9 remains lower compared to that of countries, as illustrated in Figure~\ref{fig:effect-to-logits}(a). 
At layer 10, the probability of $Y$ surpasses that of $X$ for the first time, indicating the primary occurrence of function application.
In this section, we study what happens within layer 10.

\begin{figure}
  \centering
  
  \begin{minipage}{0.35\textwidth}
    \hspace{1cm}\begin{subfigure}[t]{\textwidth}
    \centering
    \renewcommand{\arraystretch}{1.5}
      \caption{}
    \resizebox{!}{0.28\linewidth}{
      \begin{tabular}{ccc}
      \hline
     \diagbox{Vec.}{Token}  & $X$ & $Y$ \\
      \hline
     $\textbf{r}^{10}_{\text{pre}}$ & 46\% & 20\% \\
      $\sum_{h} \textbf{a}^{10, h}$ & 23\% & 21\% \\
      $\textbf{r}^{10}_{\text{mid}}$ & 54\% & 28\%\\
      $\textbf{m}^{10}$ & 0\% & 0\% \\
      $\textbf{r}^{10}_{\text{post}}$ & 29\% & 32\% \\
      \hline
    \end{tabular}}
    \end{subfigure}
    
    \vspace{\baselineskip} 
    
    \hspace{1cm}\begin{subfigure}[b]{\textwidth}
    \centering
    \renewcommand{\arraystretch}{1.3}
      \caption{}
    \resizebox{!}{0.19\linewidth}{
      \begin{tabular}{ccc}
      \hline
     \diagbox{Vec.1}{Vec.2}  & $\textbf{W}_{U}[X]$ & $\textbf{W}_{U}[Y]$ \\
      \hline
      $\textbf{m}^{10}$ & 0.048 & 0.034 \\
      $\textbf{r}^{10}_{\text{mid}}$ & 0.213 & 0.197 \\
      $\textbf{r}^{10}_{\text{post}}$ & 0.160 & 0.155 \\
      \hline
    \end{tabular}}
    \end{subfigure}
    \label{tab:1}
  \end{minipage}
  \hfill
  \begin{minipage}{0.55\textwidth}
    \centering
    \includegraphics[width=0.52\linewidth]{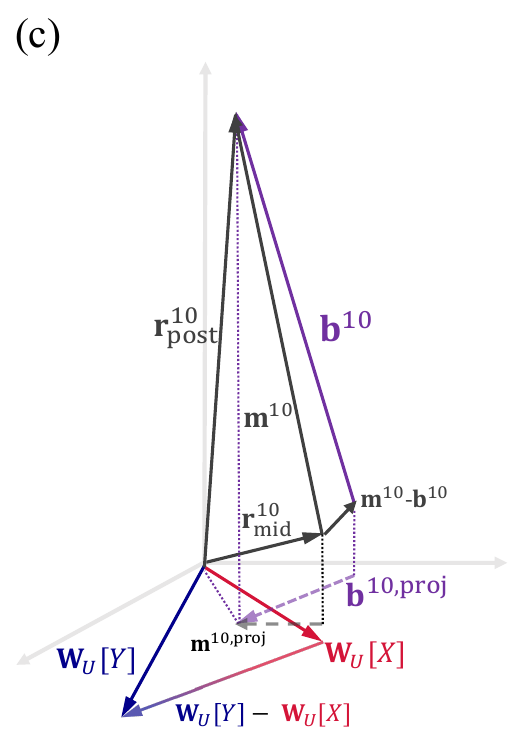}
  \end{minipage}
  \caption{
(a) The probabilities of $X$ and $Y$ decoded from various vectors. 
(b) Cosine similarity between the vectors. Note that although the cosine similarity between $\textbf{r}^{10}_{\text{post}}$ and $\textbf{W}_{U}[X]$ remains higher than that with $\textbf{W}_{U}[Y]$, when taking into account the norm, the logit value of $Y$ is higher.
(c) A simplified example illustrating the relationships among vectors in 3D space. As MLP10 generates a component $\textbf{b}^{10,\text{proj}}$ that aligns with $\textbf{W}_{U}[Y] - \textbf{W}_{U}[X]$ in direction, adding $\textbf{b}^{10,\text{proj}}$ to the residual stream achieves the ``function application,'' causing the probability of $Y$ to surpass that of $X$ for the first time during the forward pass.
}
\label{fig:projection}
\vspace{-4mm}
\end{figure}

We apply the $\texttt{Unembed}(\cdot)$ function to the $\textbf{r}^{10}_{\text{pre}}$, $\textbf{r}^{10}_{\text{mid}}$, and $\textbf{r}^{10}_{\text{post}}$, sequentially, yielding the probabilities of $X$ and $Y$ at the final positions of each node.
Results are shown in Figure~\ref{fig:projection}(a).
Notably, up to $\textbf{r}^{10}_{\text{mid}}$, the probability of capitals remains lower than that of countries, with this trend reversing at $\textbf{r}^{10}_{\text{post}}$. 
Therefore, MLP10 must write information related to the correct answers into the residual stream.
However, upon decoding $\textbf{m}^{10}$, the outputs of MLP10, we observe that the probabilities of countries and cities are both 0, indicating that $\textbf{m}^{10}$ is almost orthogonal to both $X$ and $Y$'s unembedding vectors. 
How does this contribute to enhancing the correct prediction?

To answer this question, we analyzed the cosine similarities among several pairs of vectors, and the results are reported in Figure~\ref{fig:projection}(b). 
In Figure~\ref{fig:projection}(c), we offer a 3D toy example to grasp the relationships between these vectors.
We hypothesize that despite $\textbf{m}^{10}$ being almost perpendicular to both $\textbf{W}_{U}[X]$ and $\textbf{W}_{U}[Y]$, its addition to $\textbf{r}^{10}_{\text{mid}}$ could divert the residual stream away from $\textbf{W}_{U}[X]$ and towards $\textbf{W}_{U}[Y]$, resulting in a higher probability of $Y$ in $\textbf{r}^{10}_{\text{post}}$.

To validate the above hypothesis, we project $\textbf{m}^{10}$ onto the hyperplane spanned by $\textbf{W}_{U}[Y]$ and $\textbf{W}_{U}[X]$.
Specifically, we stack $\textbf{W}_{U}[Y]$ and $\textbf{W}_{U}[X]$ as columns, constructing the corresponding projection matrix $\textbf{P}$. 
The projection of $\textbf{m}^{10}$ onto the hyperplane is achieved by its multiplication with $\textbf{P}$:
\begin{equation}
    \begin{aligned}
        & \textbf{M} = [\textbf{W}_{U}[X], \textbf{W}_U[Y]],\\
        & \textbf{P} = \textbf{M} (\textbf{M}^{\top} \textbf{M})^{-1} \textbf{M}^{\top}, \\
        & \textbf{m}^{10, \text{proj}} = \textbf{P} \textbf{m}^{10}.
    \end{aligned}
\end{equation}
If $\textbf{m}^{10, \text{proj}}$ exhibits a significant cosine similarity with the vector ($\textbf{W}_{U}[Y]$ - $\textbf{W}_{U}[X]$), then our hypothesis is verified. 
We observed that this cosine similarity is 0.361, which, while not exceptionally high, still provides insight.
Considering that $\textbf{m}^{l}$ can be decomposed into the activations of attention heads and an intercept $\textbf{b}^{l}$ designed to capture newly added influences imposed by the MLP, we further project the intercept $\textbf{b}^{10}$ to the hyperplane.
It is noteworthy that $\textbf{b}^{10,\text{proj}}$ exhibits a cosine similarity of 0.946 with ($\textbf{W}_{U}[Y]$ - $\textbf{W}_{U}[X]$).
We also analyze the GPT-2 large model and present its probability dynamics in Figure~\ref{fig:gpt2-medium-large}.
It becomes apparent that beginning from layer 27, there is a significant increase in the probability gap between $X$ and $Y$, and we measured a cosine similarity of 0.821 between $\textbf{b}^{27, \text{proj}}_{\text{large}}$ and ($\textbf{W}_{U}[Y] - \textbf{W}_{U}[X]$).

These results highlight several key points: (1) MLPs apply ``implicit functions'' by adding a task-aware component into the residual stream, guiding the residual stream towards its expected direction.
(2) The validity of the linear regression method we propose for analyzing the MLPs has once again been affirmed empirically.

We should clarify that we do not assert that the intercept invariably directs the residual stream from ``$X$'' towards ``$Y$'' in deep layers.
Understanding the underlying mechanism of specific MLP layers calls for individual scrutiny in each case, considering the potentially intricate behaviors that influence the probability of specific tokens.
We will discuss this in Appendix~\ref{apx:opt}.

Additionally, given the false negative results when directly applying the logit lens technique to $\textbf{m}^{10}$, we make two recommendations for future works.
Firstly, it is important to be cautious when applying the logit lens to MLPs or attention heads' outputs.
Secondly, when examining the outputs of modules within middle layers, it is essential to consider cosine similarity because subtle nuances may undergo significant amplification in probability after softmax.

To sum up, we have addressed the research questions listed earlier.

\section{The universal anti-overconfidence mechanisms at the final layer}
\label{sec:few-shot}

Readers may notice another phase termed ``saturation'' in the \textit{one-shot} scenario explored by Merullo et al. \cite{merullo2023mechanism}, following ``function application,'' wherein $Y$ maintains its top rank in the final layers.
In contrast, as depicted in Figure~\ref{fig:effect-to-logits}(a) in the zero-shot scenario, the probabilities of both $X$ and $Y$ drastically drop in the last two layers.
To investigate this disparity, we extended our examination from zero-shot to few-shot settings, and uncovered several intriguing findings. 
Due to page constraints, we briefly introduce these findings and reserve detailed experiments and discussions for Appendix~\ref{apx:few-shot}.

\begin{figure}[t]
    \centering
    \includegraphics[width=0.9\linewidth]{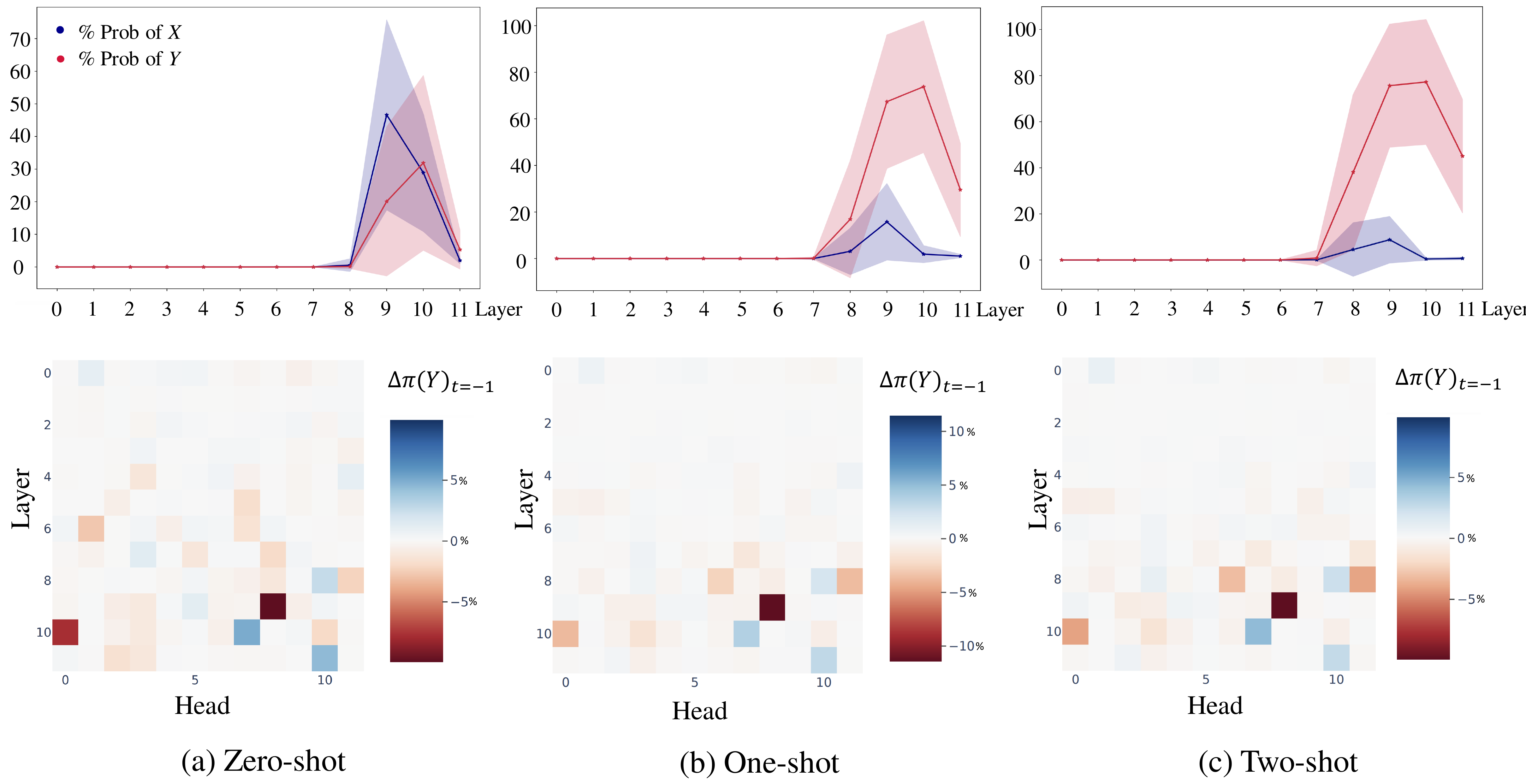}
    \caption{The sub-figures illustrate the probability dynamics of $X$ and $Y$, alongside influential heads impacting final logits across zero, one, and two-shot settings. 
    The fundamental mechanisms detected in the zero-shot scenario still work in few-shot settings.}
    \label{fig:few-shot}
    \vspace{-5mm}
\end{figure}

\paragraph{Zero-shot mechanisms still work in few-shot scenarios. (Appendix~\ref{apx:0shot-fewshot})}
Figure~\ref{fig:few-shot} shows that the identified influential heads remain primarily consistent regardless of the number of demonstrations provided. 
Additionally, the probability dynamics of $X$ and $Y$ resemble those in zero-shot scenarios: the formation of arguments and function application are clear, and in the final layers, the probabilities of both $X$ and $Y$ decrease. 
It becomes evident that the identification of the “saturation” phase by Merullo et al.~\cite{merullo2023mechanism} is not due to in-context demonstrations but rather stems from the metric they utilize, the token rank rather than detailed probability. Consequently, the final declines are overlooked.

\paragraph{ICL improves model confidence to generate capital's \textit{name}. (Appendix~\ref{apx:icl})}
We discovered that with more ICL demonstrations, the emergence of probabilities for $X$ and $Y$ occurs earlier, for instance, in layer 8 in few-shot scenarios compared to layer 9 in zero-shot scenarios (see top figures in Figure~\ref{fig:few-shot}).
Meanwhile, the decline in the final layer is relatively slight. 
We found that in zero-shot scenarios, the model tends to output ``safe'' tokens such as ``not'' following ``The capital of X is.'' 
In few-shot scenarios, we detected some potential mechanisms that prompt the model to recognize the output domain to be the name of the capital, thereby enhancing the confidence in outputting $Y$.


\paragraph{The final layer avoids overconfidence. (Appendix~\ref{apx:overconfidence})} 
We revealed that the final layer of language models takes two steps to avoid outputting overconfident answers.

\begin{wrapfigure}{r}{0.36\linewidth}
    \centering
    \includegraphics[width=\linewidth]{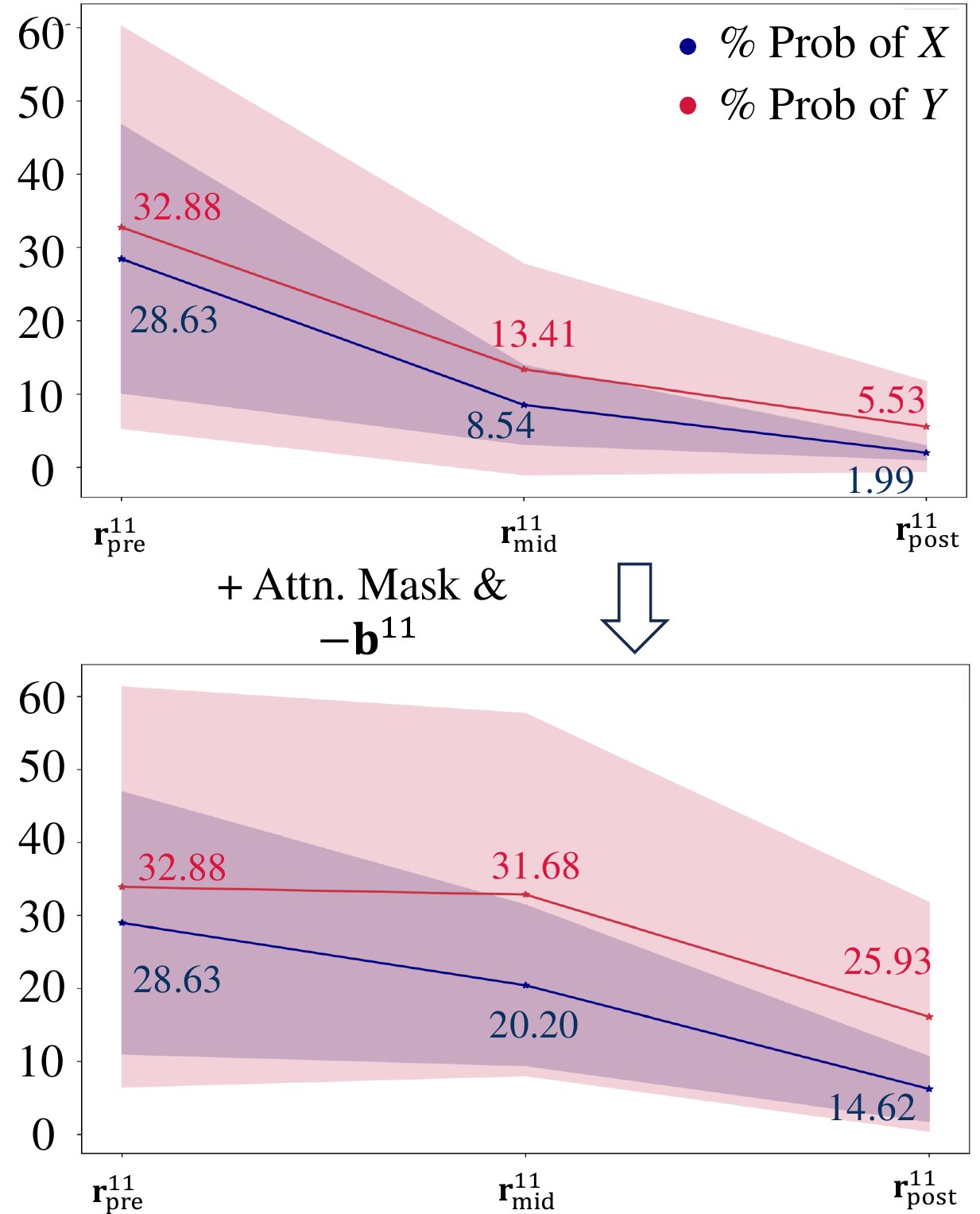}
    \caption{The anti-overconfidence mechanism at the last layer imposed by attention heads and MLPs.
    This mechanism can be suppressed according to our analysis.}
    \label{fig:main-suppress-fig}
    \vspace{-3mm}
\end{wrapfigure}

(1) Attention heads incorporate frequent tokens into the residual stream.
We found that most attention heads in the last layer focus on the first position (see Figure~\ref{fig:layer11-attn}) where their OV circuits output frequent English tokens such as "and", "a", and ".", etc.
These frequent tokens are copied from the first position to the final position, diluting the probability of $Y$.

(2) The MLP generates a vector that steers the residual stream towards an ``average'' token, the frequency-weighted average token embedding over the training corpus, denoted as $\mathbb{E}[\textbf{W}_{U}]$.
When applying our linear regression analysis to the final MLP layer and obtaining the corresponding intercept term $\textbf{b}$, there is high similarity between ($\mathbb{E}[\textbf{W}_{U}]$ - $\textbf{W}_{U}[Y]$) and $\textbf{b}$, indicating that this MLP directs the residual stream from the correct answer $Y$ to the ``average'' token. 

These two steps are intuitive: the model incurs a large loss when a confident answer is incorrect. 
By aligning the model outputs with the expected logits distribution across the training corpus, the model losses could be effectively minimized.
They can be suppressed for a confident output by (1) applying attention masks to block information flow from the first position and (2) subtracting $\textbf{b}$ from the residual stream. 
For example, as shown in Figure~\ref{fig:main-suppress-fig}, on GPT-2 small, we improved the zero-shot output probability of $Y$ from 5.53\% to 25.93\%. 

\section{Conclusion}
In this paper, we conduct a thorough investigation into the factual recall mechanisms utilized by language models. 
We studied the mechanisms behind ``argument passing'' and ``function application.''
Furthermore, we uncovered the prevalence of anti-overconfidence mechanisms in language models.
Our proposed analysis method, based on linear regression, effectively decomposes MLP outputs into components that are easily understandable to humans. 
This method has been substantiated through numerous empirical experiments and lays a valuable foundation for our interpretations.
Finally, we study the anti-overconfidence mechanisms in the final layer, which is a universal mechanism regardless of models, tasks, and the number of in-context demonstrations.

\begin{ack}
We appreciate the valuable discussions with Zhuocheng Gong (Peking University) and Wei Yao (Renmin University of China), assistance in figure presentation from Tao Tan (Renmin University of China), and writing suggestions from Shuqi Li (Renmin University of China).
This study is sponsored by CCF-BaiChuan-Eblech Foundation Model Fund.
\end{ack}

\bibliography{good_luck}
\bibliographystyle{plain}

\appendix

\section{The universal anti-overconfidence mechanisms at the final layer}
\label{apx:few-shot}

Readers may notice another phase termed ``saturation'' in the \textit{one-shot} scenario explored by Merullo et al. \cite{merullo2023mechanism}, following ``function application,'' wherein $Y$ maintains its top rank in the final layers.
In contrast, as depicted in Figure~\ref{fig:effect-to-logits}(a) in our zero-shot scenario, the probabilities of both $X$ and $Y$ drastically drop in the last two layers.
We then extend our investigation from the zero-shot to few-shot settings to delve deeper.

\subsection{Zero-shot mechanisms still work in few-shot scenarios}
\label{apx:0shot-fewshot}
We construct two new datasets: one with a single in-context demonstration (one-shot) and another with two (two-shot). 
We achieve this by appending one or two prefixes in the form of ``The capital of $X_{\text{ICL}}$ is $Y_{\text{ICL}}$,'' respectively, to the original prompts. 
each ($X_{\text{ICL}}$, $Y_{\text{ICL}}$) is randomly selected and distinct from ($X$, $Y$).

Using these datasets, we conduct ``patching $\textbf{a}^{l,h}_{t=-1} \rightarrow \textbf{r}^{11}_{\text{post}, t=-1}$'' experiments as described in \S\ref{sec:attention-pass-arguments}. 
We also plot the probability dynamics of $X$ and $Y$ across the layers. 
The results are shown in Figure \ref{fig:few-shot}.
Upon comparing the results of 0-shot and few-shot experiments, we observe that the mechanisms studied in the zero-shot scenario persist in the few-shot scenarios.
This persistence manifests in two aspects: Firstly, the consistency of probability dynamics and influential heads across varying numbers of ICL demonstrations; Secondly, the consistent decline in the probability of both $X$ and $Y$ as they reach the final layers across all scenarios.
Now, it becomes evident that the identification of the ``saturation'' phase by Merullo et al. \cite{merullo2023mechanism} is not due to in-context demonstrations but rather stems from the metric they utilize, the token rank rather than detailed probability. 
Consequently, the final declines are overlooked.
The following subsections will delve into the reasons behind this universal decline at the final layer.

\subsection{ICL improves model confidence in generating capitals' \textit{name}}
\label{apx:icl}
Before delving into the universal decline, there is an extra crucial finding when comparing the zero-shot results to the few-shot results.
One noticeable distinction lies in the emergence of probabilities for $X$ and $Y$ at layer 8 in few-shot settings, compared to their emergence until layer 9 in 0-shot settings. 
This distinction can be attributed to L8H11 and L8H6, where L8H11 is identified as a ``mover'' head.
As the number of ICL demonstrations increases, these heads become more influential.

The ICL demonstrations enhance the model's confidence in generating the \textit{name} of capitals from two key aspects:
Firstly, an additional mover head, L8H11, moves the country names to the final position, implying that with more ICL demonstrations, the model gains insights into the ``argument'' earlier, allowing for deeper processing of the answer across more layers.
Secondly, in the zero-shot scenario, the model tends to produce ``safe'' tokens such as ``not'' to follow the prompt, resulting in sentences like ``The capital of $X$ is not...'' which, although coherent, lack useful information.
This might stem from a lack of confidence in the output domain or format.
However, in few-shot settings, L8H6 pays considerable attention from the final token to $Y_{\text{ICL}}$, i.e., the in-context capital names.
It seems prompt the model to recognize that the expected answer is the \textit{name} of a capital city.
We leave the exploration of the following questions to future research, as the answer is non-trivial: How do ICL demonstrations trigger the earlier activation of mover heads? How does the model determine that it should output city names?

\subsection{Negative heads suppress \texorpdfstring{$X$}{}}
\begin{wrapfigure}{r}{0.32\textwidth}
\vspace{-8mm}
\begin{center}
    \includegraphics[width=0.32\textwidth]{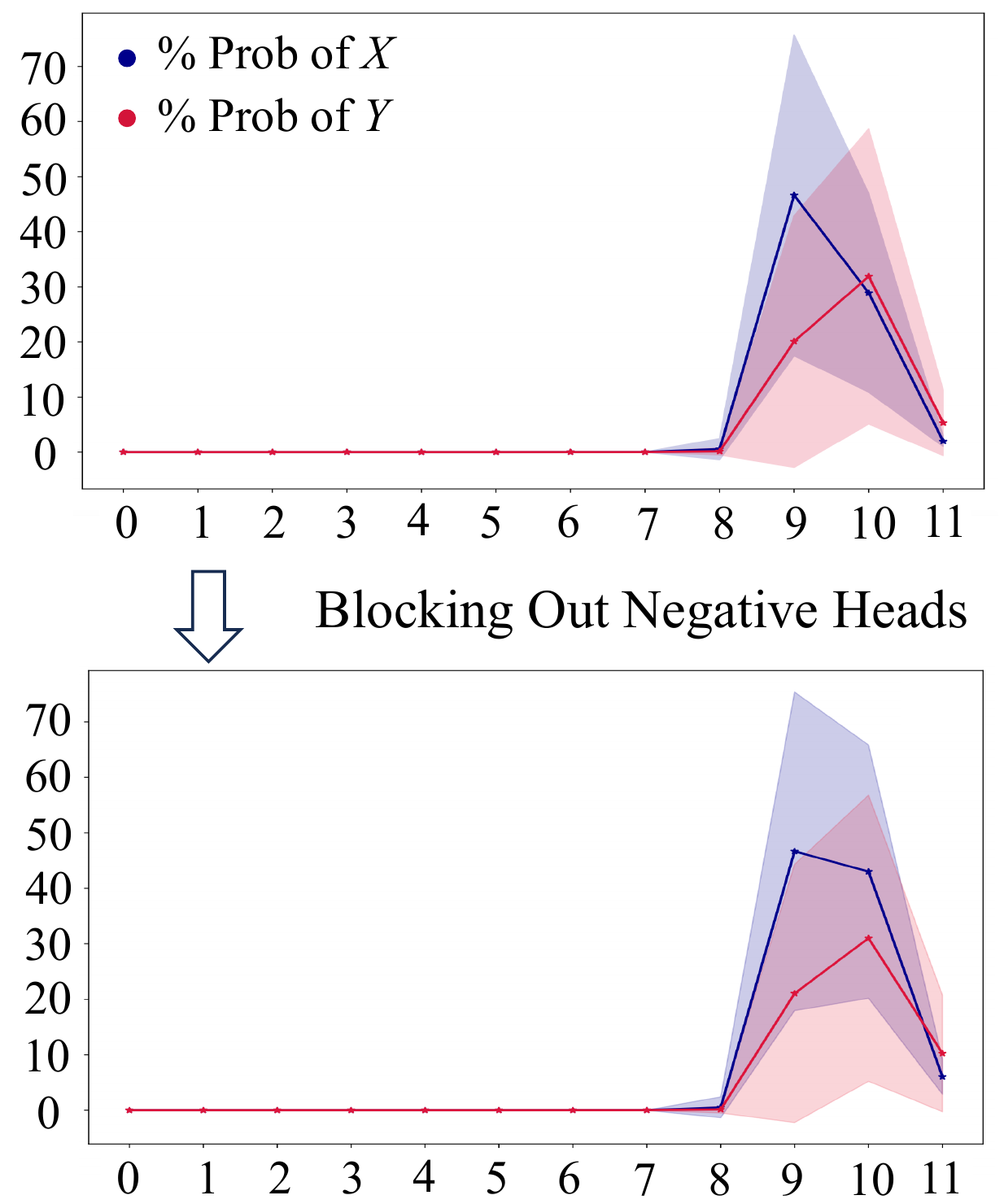}
    \end{center}
    \caption{When blocking out the negative heads in zero-shot settings, only suppression of $X$ is slightly alleviated.}
    \label{fig:block-negative}
\end{wrapfigure}

As shown in Figure~\ref{fig:few-shot}, we identify two negative heads, L10H7 and L11H10, which suppress the correct prediction, irrespective of the number of shots.
Take the zero-shot scenario as an example.
We investigate the impact of these heads.
We patch their outputs with their average activations given the set $\{p_{\text{intervention},n \leq N}\}$ at position $X$, thereby eliminating the country identity in their outputs and thus blocking them out.
We can gain insight into their suppression effect by examining the probabilities of $X$ and $Y$ after blocking out these heads. 
The zero-shot probability dynamics when patching these heads are illustrated in Figure~\ref{fig:block-negative}.
Even though blocking these negative heads mitigates the suppression of $X$ in layer 10, it remains strong in layer 11. 
The probability of $Y$ experiences a slight increase but remains relatively stable and low overall. 
These findings suggest that both negative heads primarily suppress the probability of $X$. 
This observation is intriguing, especially considering that, despite L11H10 and L10H7 being the only two negative heads in the last layers, they do not appear to suppress $Y$ directly. 
What mechanism, then, is at play in decreasing the probability of $Y$ at the final layer?

\subsection{The final layer avoids overconfidence}
\label{apx:overconfidence}
\begin{figure}[t]
    \centering
    \includegraphics[width=\linewidth]{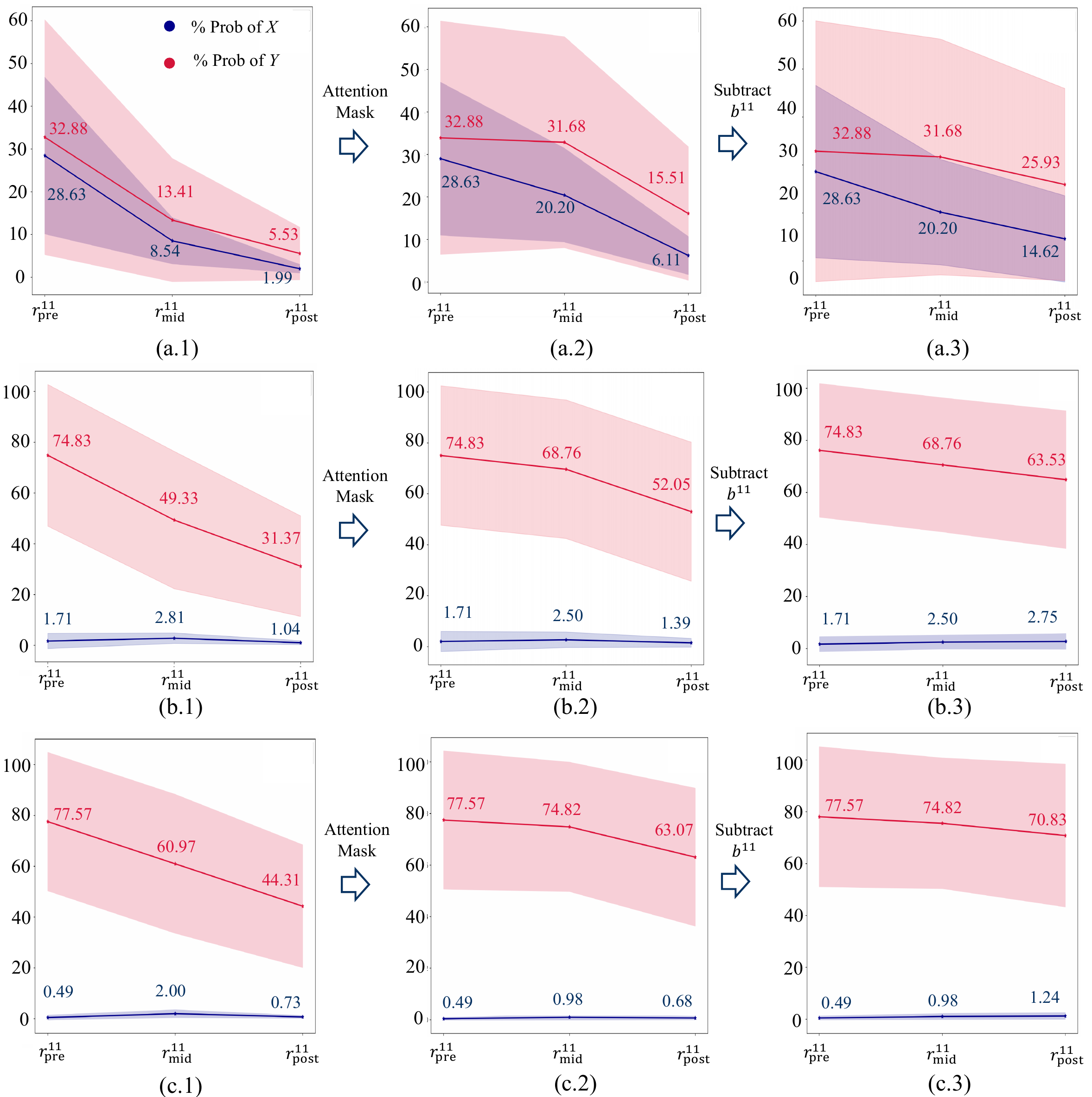}
    \caption{
    Given the country-capital task, the probability of $X$ and $Y$ in the final layer of GPT-2 small in (a) zero-shot, (b) one-shot, and (c) two-shot scenarios.
    The suppression of $Y$ by the attention heads in layer 11 can be eliminated by reallocating their attention. 
    The suppression of $Y$ by MLP11 can be mitigated by subtracting $\textbf{b}^{11}$ from the residual stream. }
    \label{fig:gpt2-suppress}
\end{figure}

Within layer 11, as depicted in Figure~\ref{fig:gpt2-suppress}(a.1), (b.1), and (c.1), we observe a consecutive decline in the probability of $Y$ at $\textbf{r}^{11}_{\text{mid}}$ and $\textbf{r}^{11}_{\text{post}}$, regardless of the number of demonstrations.
This suggests that certain attention heads in this layer, along with MLP11, contribute to the suppression of $Y$.
This phenomenon is commonly observed across various models and tasks. 
We suggest that the behavior observed could be interpreted as an anti-overconfidence strategy employed within the final layer. 
This strategy could mitigate the risk of large losses stemming from an incorrect yet confident prediction.

\begin{figure}[t]
    \centering
    \includegraphics[width=\linewidth]{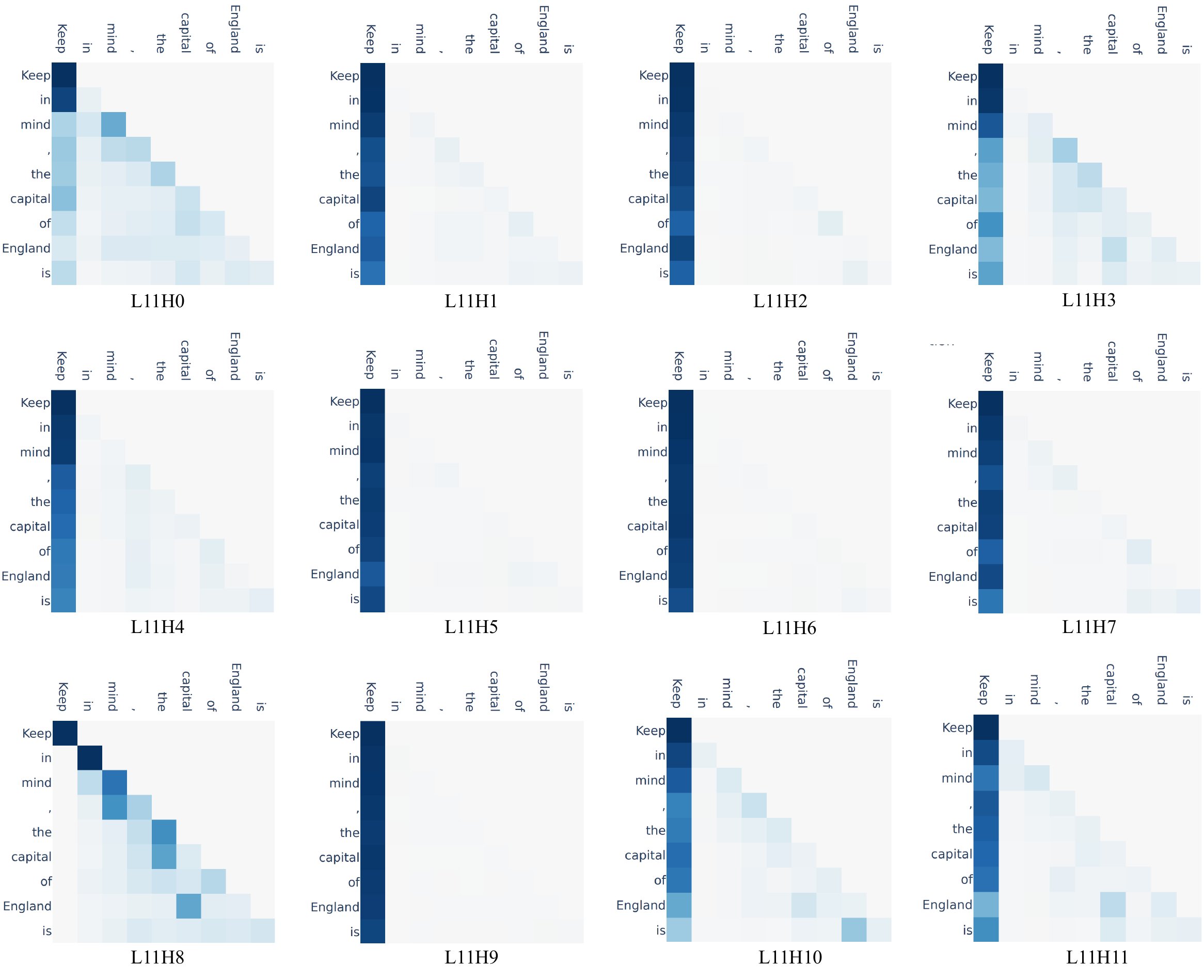}
    \caption{The attention patterns in layer 11 of GPT-2 small in the country-capital task. }
    \label{fig:layer11-attn}
\end{figure}

\paragraph{Attention heads incorporate frequent tokens into the residual stream}
Despite our efforts to block out each attention head, the suppression effect persists, suggesting that a large portion of all heads may contribute together to the suppression of $Y$. 
Visualizing the attention patterns of each head (see Figure~\ref{fig:layer11-attn}) provides more insights. 
It becomes evident from the figure that most heads ``park'' their attention on the initial token. 
This pattern is consistent across various tasks and models, indicating its generality. 
Previous works~\cite{softmaxbyone,xiao2023efficient} suggest that this attention pattern indicates that attention heads are inactive and indifferent to the context.
We analyzed the output of the OV circuit in these heads at the initial position and observed that some of them output frequent English tokens such as "the," ",", ".", "a," and "and," while others produce letters and some subwords irrelevant to the task. 
We hypothesize that these inactive heads move information in the initial position to the final position, diluting the content of $Y$ in the residual stream. 
To test our hypothesis, we applied attention masks to restrict these heads to attend solely to the final position (i.e., only the diagonal is unmasked).
The result, as depicted in Figure~\ref{fig:gpt2-suppress}(a.2), (b.2), and (c.2), shows a mitigation of suppression at $\textbf{r}^{11}_{\text{mid}}$.
For example, with zero-shot, We improved the probability of $Y$ at $\textbf{r}^{11}_{\text{mid}}$ from 13.41\% to 31.68\%.
\begin{table}[h]
\small
    \centering
    \renewcommand{\arraystretch}{1.3}
    \caption{Learned coefficients for layer 11.}
    \resizebox{0.9\linewidth}{!}{
    \begin{tabular}{ccccccccccccc}
    \toprule
        $w^{11,0}$ & $w^{11,1}$ & $w^{11,2}$ & $w^{11,3}$ & $w^{11,4}$ & $w^{11,5}$ & $w^{11,6}$ & $w^{11,7}$ & $w^{11,8}$ & $w^{11,9}$ & $w^{11,10}$ & $w^{11,11}$ & $w^{11,r}$\\\hline
         -0.22 & 0.74 & -0.11 & -0.31 & -0.05 & 1.55 & 0.54 & 0.42 & 0.13 & 0.98 & 0.13 & 0.22 & 0.01\\\bottomrule
    \end{tabular}}
    \label{tab:lr2}
\end{table}

\paragraph{The final MLP prefers ``safe'' prediction}
We examined the regression coefficients of Eq.~\ref{eq:lr} for the layer 11, as detailed in Table~\ref{tab:lr2}. 
It is observed that most attention heads are assigned with positive weights, some even surpassing 1. 
Meanwhile, $\textbf{r}^{11}_{\text{pre}}$, which contains much information of $Y$, has a relatively modest weight.
These findings suggest that MLP11 further amplifies the suppression of $Y$ imposed by attention heads.

Further, we delve into examining the learned intercept of the final layer to gain deeper insights into the workings of the final MLP.
We hypothesize that the final MLP aligns the residual stream with $\mathbb{E}[\textbf{W}_{U}]$, the expectation of output logits over the training corpus, thereby effectively mitigating significant training losses.
$\mathbb{E}[\textbf{W}_{U}]$ is computed as the average of $w_{v} \cdot \textbf{W}_{U}[v]$ over $V$ tokens. 
Here, $w_v$ represents the relative word frequency derived from OpenWebText~\cite{Gokaslan2019OpenWeb}, the training corpus of GPT-2. 
We utilize the word frequency data provided by~\cite{kobayashi-etal-2023-transformer}.

In the case of GPT-2 medium and large models, which consist of 24 and 36 layers, respectively, when we project the individual intercepts \( \textbf{b}^{23}_{\text{medium}} \) and \( \textbf{b}^{35}_{\text{large}} \) onto the hyperplane spanned by \( \mathbb{E}[\textbf{W}_{U}] \) and \( \textbf{W}_{U}[Y] \), we observe that the projected intercepts point from \( \textbf{W}_{U}[Y] \) towards \( \mathbb{E}[\textbf{W}_{U}] \). 
This directional alignment is supported by their respective cosine similarities, measuring 0.976 and 0.661 with respect to the vector (\( \mathbb{E}[\textbf{W}_{U}] - \textbf{W}_{U}[Y] \)).
In GPT-2 Small, a notable correlation is evident between \( \textbf{b}^{11} \) and \( \mathbb{E}[\textbf{W}_{U}] \), indicated by a cosine similarity of 0.933, while the similarity between $\textbf{b}^{11, \text{proj}}$ and ($\mathbb{E}[\textbf{W}_{U}] - \textbf{W}_{U}[Y]$) remains relatively small.

We acknowledge that the hypothesis regarding the final MLP layer remains anecdotal since we have only tested it within the GPT-2 family due to a lack of training corpus data for other models.
Nevertheless, subtracting the intercept from the residual stream in the final layer empirically proves to be an effective approach in mitigating the suppression of correct predictions. 
Results on GPT-2 small are depicted in Figure~\ref{fig:gpt2-suppress}(a.3), (b.3), and (c.3).
Through this approach, for example, we were able to enhance the zero-shot probability of $Y$ at $\textbf{r}^{11}_{\text{post}}$ in GPT-2 small from $15.51\%$ to $25.93\%$.
The efficacy of two anti-suppression techniques—namely, applying attention masks and subtracting intercepts—has been validated across various models and tasks. 
Please refer to Appendix \ref{apx:more-task}, \ref{apx:opt} and \ref{apx:llama} for further details.

\section{Related works}
In investigating the factual recall of language models, Geva et al. \cite{geva-etal-2023-dissecting,geva2022transformer} traced the information change in factual recall tasks.
Yu et al. \cite{yu2023characterizing} and Ortu et al. \cite{ortu2024competition} studied important attention heads in a counterfactual setting.
\cite{heinzerling2024monotonic} is a contemporary work that finds the specific direction in activation space encodes the semantics of model final outputs. 
None of these works provide an in-depth analysis of the mechanisms behind each module's function.
Merullo et al. \cite{merullo2023mechanism} described the factual recall as a process of ``passing an argument and then applying the function.'' 
Based on their analogy, we take a further step to provide the details of how ``arguments'' are identified from the context and passed to the ``function.''
Similar to our paper, Geva et al. \cite{geva-etal-2023-dissecting} also found that the OV matrix of attention heads contains knowledge.
Other related works focus on factual recall and mechanistic interpretability.
Our findings do not overlap with theirs but complement each other.
Lieberum et al. \cite{lieberum2023does} conducted a multiple-choice task involving factual knowledge within large language models. 
Yu et al. \cite{yu2024locating} pinpointed the location of knowledge within neurons in MLPs and interpreted how the residual stream activates knowledge in MLPs. 

Elhage et al. \cite{elhage2021mathematical} proposed that Transformers operate using a ``memory management'' approach, where modules are reading from and writing to the residual stream.
Dao et al. \cite{dao2023adversarial} trained a 4-layer model and found that certain attention heads could erase the outputs of previous attention heads.
McGrath et al. \cite{mcgrath2023hydra} observed the erasure of tokens by MLPs.
More than observations, we dissected the role of MLPs by \textit{quantifying} their impact on individual heads, and interpreting their role in ``function application.''

\section{Details of datasets}
\label{apx:dataset}
\paragraph{The country-capital task}
Table~\ref{tab:prompt-templates-country} shows prompt templates we used.
Here are country-capital ($X$, $Y$) pairs we used: (China, Beijing), (USA, Washington. D.C.), (Russia, Moscow), (England, London), (France, Paris), (Japan, Tokyo), (Italy, Rome), (Canada, Ottawa), (Australia, Canberra), (Spain, Madrid), (Egypt, Cairo), (Portugal, Lisbon), (Austria, Vienna), (Greece, Athens), (Thailand, Bangkok).

\paragraph{The product-developer task}
In addition to the country-capital task, we also conduct experiments on a product-developer task.
Table~\ref{tab:prompt-templates-product} shows prompt templates we used.
Here are Product-Developer ($X$, $Y$) pairs we used: (iPhone, Apple), (Windows7, Microsoft), (GTX1060, Nvidia), (YouTube, Google), (Firefox, Mozilla), (VirtualBox, Oracle), (Instagram, Facebook), (Pentium, Intel), (Steam, Valve), (Radeon, AMD), (Photoshop, Adobe), (PlayStation, Sony), (Kindle, Amazon), (GameBoy, Nintendo). 

\begin{table}[h]
\vspace{3mm}
  \centering
\begin{minipage}{0.49\textwidth}
    \centering
    \caption{Prompt templates used in the country-capital task.}
    \resizebox{\textwidth}{!}{
    \begin{tabular}{cc}\toprule
    Index & Prompt Template\\\hline
        0 & It's crucial to know that the capital of $X$ is\\
        1 & You are right to say that the capital of $X$ is\\
       2 & Therefore, it's correct to state that the capital of $X$ is\\
      3 &When asked, always remember that the capital of $X$ is\\
    4&We confirm that the capital of $X$ is\\
    5&Don't forget, the capital of $X$ is\\
    6&Bear in mind, the capital of $X$ is\\
    7&Keep in mind, the capital of $X$ is\\
    8&Just a reminder, the capital of $X$ is\\
    9&As we all know, the capital of $X$ is\\
    10&According to the textbook, the capital of $X$ is\\
    11&I am sure that the capital of $X$ is\\
    12&Without a doubt, the capital of $X$ is\\
    13&In case you didn't know, the capital of $X$ is\\
    14&To emphasize, the capital of $X$ is\\\bottomrule
    \end{tabular}}
    \label{tab:prompt-templates-country}
  \end{minipage}
  \hfill
  \begin{minipage}{0.49\textwidth}
    \centering
    \caption{Prompt templates used in the product-developer task.}
    \resizebox{\textwidth}{!}{
    \begin{tabular}{cc}\toprule
    Index & Prompt Template\\\hline
        0 & It's crucial to know that $X$ is developed by\\
        1 & You are right to say that $X$ is developed by\\
       2 & Therefore, it's correct to state that $X$ is developed by\\
      3 &When asked, always remember that $X$ is developed by\\
    4&We confirm that $X$ is developed by\\
    5&Don't forget, $X$ is developed by\\
    6&Bear in mind, $X$ is developed by\\
    7&Keep in mind, $X$ is developed by\\
    8&Just a reminder, $X$ is developed by\\
    9&As we all know, $X$ is developed by\\
    10&According to the textbook, $X$ is developed by\\
    11&I am sure that $X$ is developed by\\
    12&Without a doubt, $X$ is developed by\\
    13&In case you didn't know, $X$ is developed by\\
    14&To emphasize, $X$ is developed by\\\bottomrule
    \end{tabular}}
    \label{tab:prompt-templates-product}
  \end{minipage}
\end{table}

\newpage
\section{The influence of prompt templates}

In experiments in the main text, the task word ``capital'' always precedes the country name $X$, as shown in Table~\ref{tab:prompt-templates-country}.
We now investigate whether altering the order of these keywords affects the factual recall.
To explore this, we construct a new set of prompts, detailed in Table~\ref{tab:prompt-templates-country-reverse}.
We have the following key findings:

\begin{table}[t] 
   \centering
    \caption{Prompt templates in the country-capital task where $X$ precedes the task word ``capital.''}
    \vspace{2mm}
    \label{tab:prompt-templates-country-reverse}  
    \resizebox{0.55\linewidth}{!}{
    \begin{tabular}{cc}\toprule
    Index & Prompt Template\\\hline
        0 & It's crucial to know that $X$'s capital is\\
        1 & You are right to say that $X$'s capital is\\
       2 & Therefore, it's correct to state that $X$'s capital is\\
      3 &When asked, always remember that $X$'s capital is\\
    4&We confirm that $X$'s capital is\\
    5&Don't forget, $X$'s capital is\\
    6&Bear in mind, $X$'s capital is\\
    7&Keep in mind, $X$'s capital is\\
    8&Just a reminder, $X$'s capital is\\
    9&As we all know, $X$'s capital is\\
    10&According to the textbook, $X$'s capital is\\
    11&I am sure that $X$'s capital is\\
    12&Without a doubt, $X$'s capital is\\
    13&In case you didn't know, $X$'s capital is\\
    14&To emphasize, $X$'s capital is\\\bottomrule
    \end{tabular}}
\end{table}

\begin{figure}[t] 
    \centering
    \includegraphics[width=0.8\linewidth]{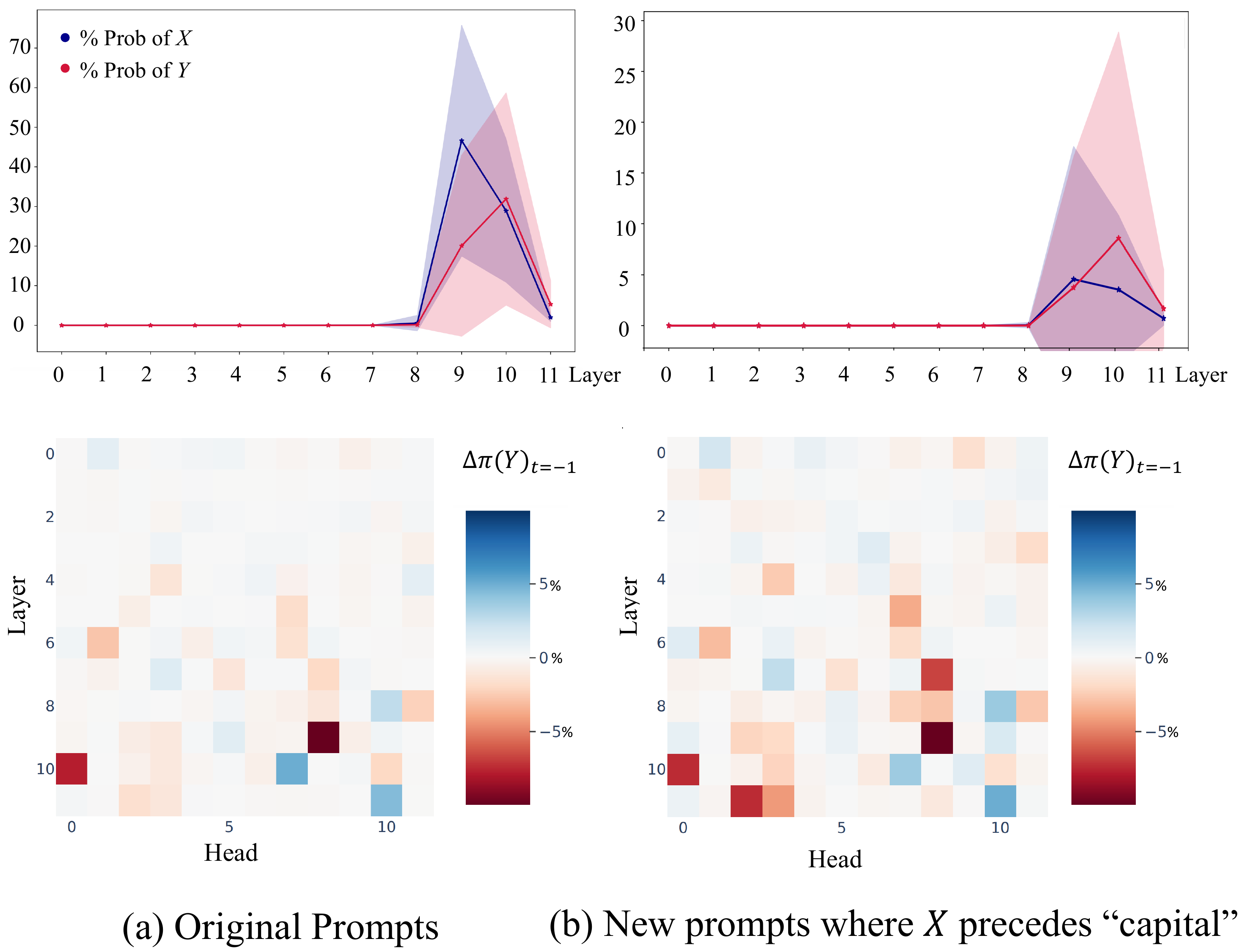}
    \caption{Probability dynamics and detected influential heads across varied prompt templates.}
    \label{fig:reverse}
\end{figure}

(1) \textbf{Consistent influential heads}:
Comparing Figure~\ref{fig:reverse}(a) and Figure~\ref{fig:reverse}(b), we can see that despite template differences, crucial task-specific heads, such as L9H8 and L10H0, maintain significance.
L10H7 and L11H10 continue to suppress the correct prediction.
Notably, some heads become more influential when using new templates such as L11H2 and L7H8.
L11H2 is a mover head, and L7H8 attends to ``capital,'' appearing to impact the formation of task semantics, which falls out of the scope of this paper.

(2) \textbf{Consistent probability dynamics' patterns}:
The probability dynamics pattern remains akin to the original when using the new templates.
Specifically, at layer 9, $X$ and $Y$ become evident in the residual stream, while at layer 10, the probability of $Y$ surpasses that of $X$ for the first time. 
The cosine similarity between current $\textbf{b}^{10, \text{proj}}$ and ($\textbf{W}_{U}[Y] - \textbf{W}_{U}[X]$) is 0.944.
Recall that this similarity stands at 0.946 when employing the original prompt templates.
The consistency across various templates underscores our previous findings regarding MLP10's function in utilizing the intercept to steer the direction of the residual stream.

\begin{figure}[t]
    \centering
    \includegraphics[width=0.9\linewidth]{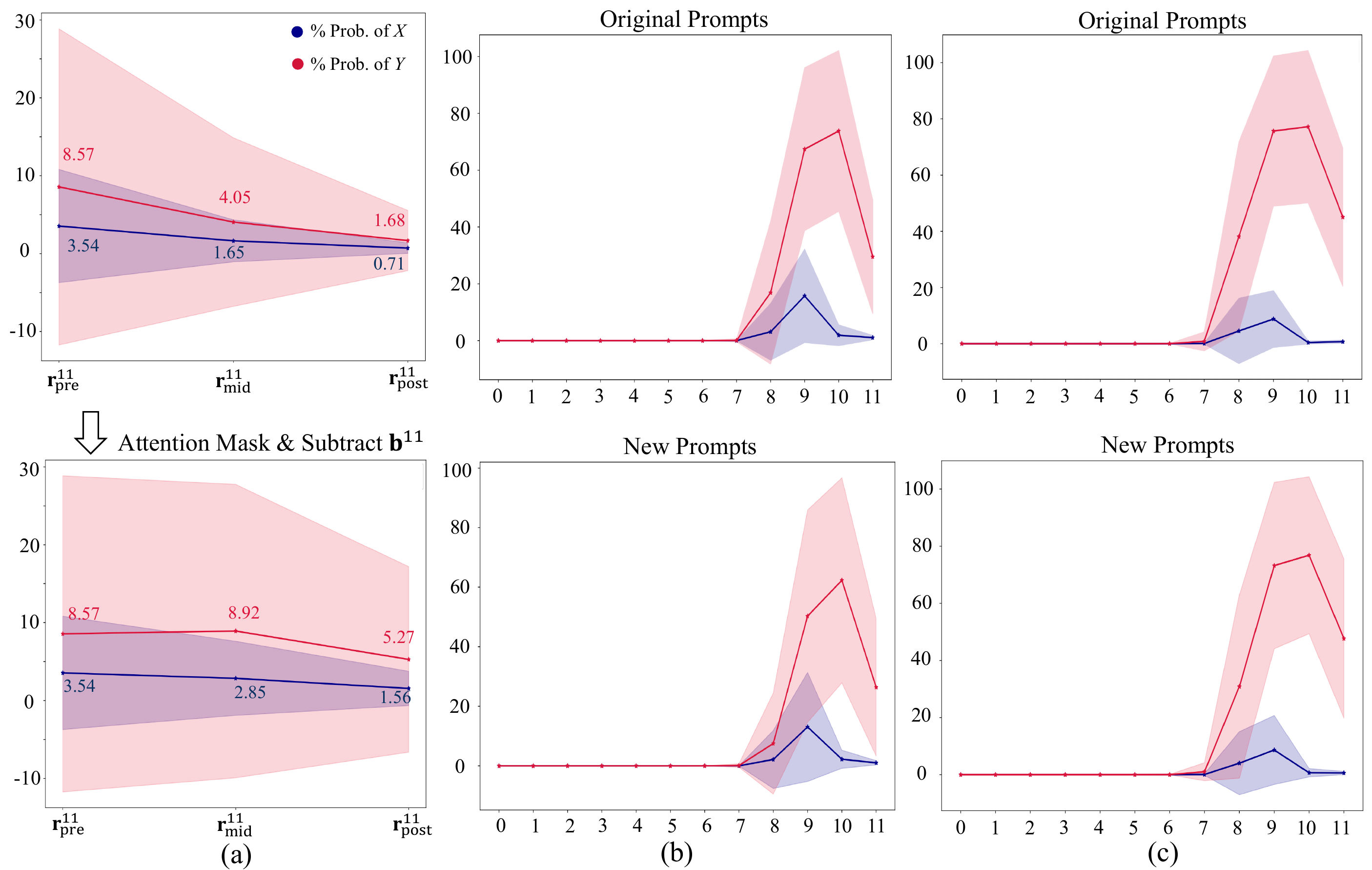}
    \caption{(a) Our proposed techniques, including applying attention masks and subtracting the intercept from the output, can suppress the anti-overconfidence mechanism in the last layer.
    (b) One-shot probability dynamics given original/new prompts.
    (c) Two-shot probability dynamics given original/new prompts.}
    \label{fig:reverse-mitigate}
\end{figure}

Moreover, at layer 11, the model exhibits the anti-overconfidence mechanism, with the cosine similarity between current $\textbf{b}^{11}$ and \( \mathbb{E}[\textbf{W}_{U}] \)  measuring at 0.89. 
This value is also notably close to 0.933 observed when employing the original templates.
As shown in Figure~\ref{fig:reverse-mitigate}(a), the anti-overconfidence mechanism can be suppressed by the two techniques we proposed.

(3) \textbf{Prompt templates impact zero-shot prediction confidence}:
The primary distinction observed when utilizing two sets of prompts resides in the probability value.
For prompts where $X$ precedes ``capital,'' the probabilities of both $X$ and $Y$ are lower.
As depicted in Figure~\ref{fig:reverse}, this divergence originates primarily from layer 9. 
At layer 9, with original prompts, the probability of $X$ could reach 50\%, whereas with the new prompts, it decreases to a mere 5\%. 

Upon scrutinizing the top decoded tokens given the new prompts, we found that ``located'' emerges as the predominant choice for output at layer 9 across almost all test samples.
However, adding prefixes ``$X_{\text{ICL}}$'s capital is $Y_{\text{ICL}}$'' before our new prompts, to create few-shot test samples, eliminates the disparity of probability dynamics between given the new prompts and the original ones.
The results are shown in Figure~\ref{fig:reverse-mitigate}(b) and (c).

It seems that our new prompts, where we place ``$X$'' before ``capital,'' exacerbate the model's uncertainty regarding the output domain, resulting in the model favoring the word ``located'' as a safe output. 
This uncertainty is effectively eliminated by in-context learning, aligning with the findings in Section~\ref{sec:few-shot} and Appendix~\ref{apx:few-shot}. 
Understanding how in-context learning boosts the model's confidence remains a complex question, which we defer to future research endeavors as mentioned in the ``Limitation'' section.

\section{Identifying the MLP layer for ``function application''}
\begin{figure}[t]
    \centering
    \includegraphics[width=0.7\textwidth]{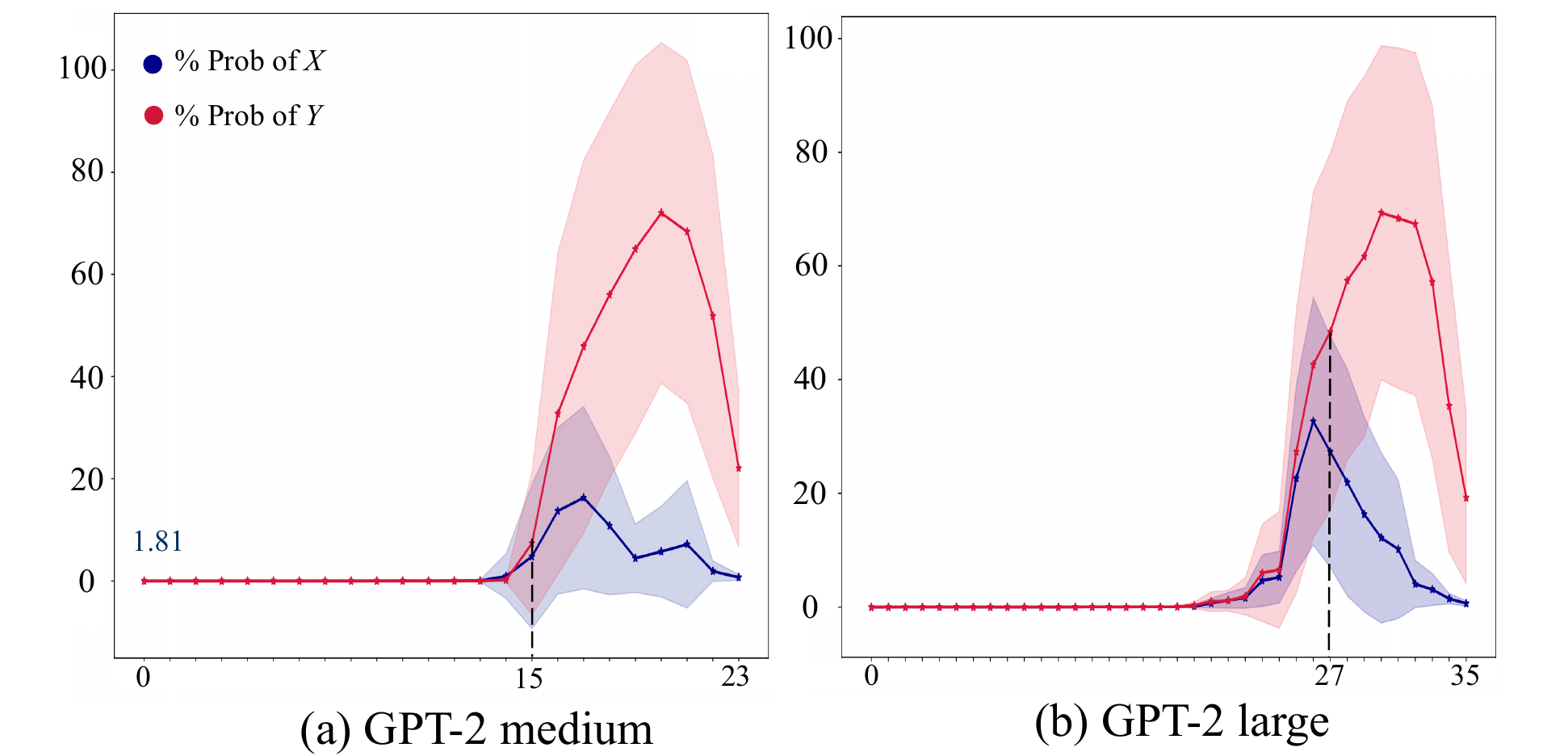}
    \caption{The probability dynamics across layers of GPT-2 medium and GPT-2 large in the zero-shot country-capital task.}
    \label{fig:gpt2-medium-large}
\end{figure}

In our main text, for simplicity's sake, we followed Merullo et al. \cite{merullo2023mechanism} to identify the layer where the probability of $Y$ surpasses that of $X$ as the point of function application.
However, this principle is not always accurate. 
A more general principle would involve examining the layers ``around'' the point where the probabilities of $X$ and $Y$ begin to diverge significantly.
To illustrate this, Figure~\ref{fig:gpt2-medium-large} showcases the zero-shot probabilities of $X$ and $Y$ across the layers in GPT-2 medium and GPT-2 large models.

The behavior of GPT-2 medium mirrors that of GPT-2 small.
The probability of $Y$ exceeds that of $X$ for the first time at layer 15.
Upon analyzing the intercepts at each layer of GPT-2 medium, we found that at layer 15, $\textbf{b}^{15,\text{proj}}$ indeed demonstrates the highest alignment with $\textbf{W}_{U}[Y] - \textbf{W}_{U}[X]$, boasting a cosine similarity of 0.661.

However, in GPT-2 large, the scenario is slightly different. 
Here, it is at layer 22 where the probability of $Y$ marginally surpasses that of $X$. 
Following this, the probabilities of $X$ and $Y$ remain comparable until layer 27, where the probability of $X$ experiences a significant decrease, coinciding with the layer where the projected intercept aligns most closely with $\textbf{W}_{U}[Y] - \textbf{W}_{U}[X]$, with the cosine probability of 0.821.
Hence, it is in layer 27, rather than layer 22, where we recognize the occurrence of function application.

\section{GPT-2 small on the ``product-developer'' task}
\label{apx:more-task}

\begin{figure}[ht]
    \centering
    \includegraphics[width=\linewidth]{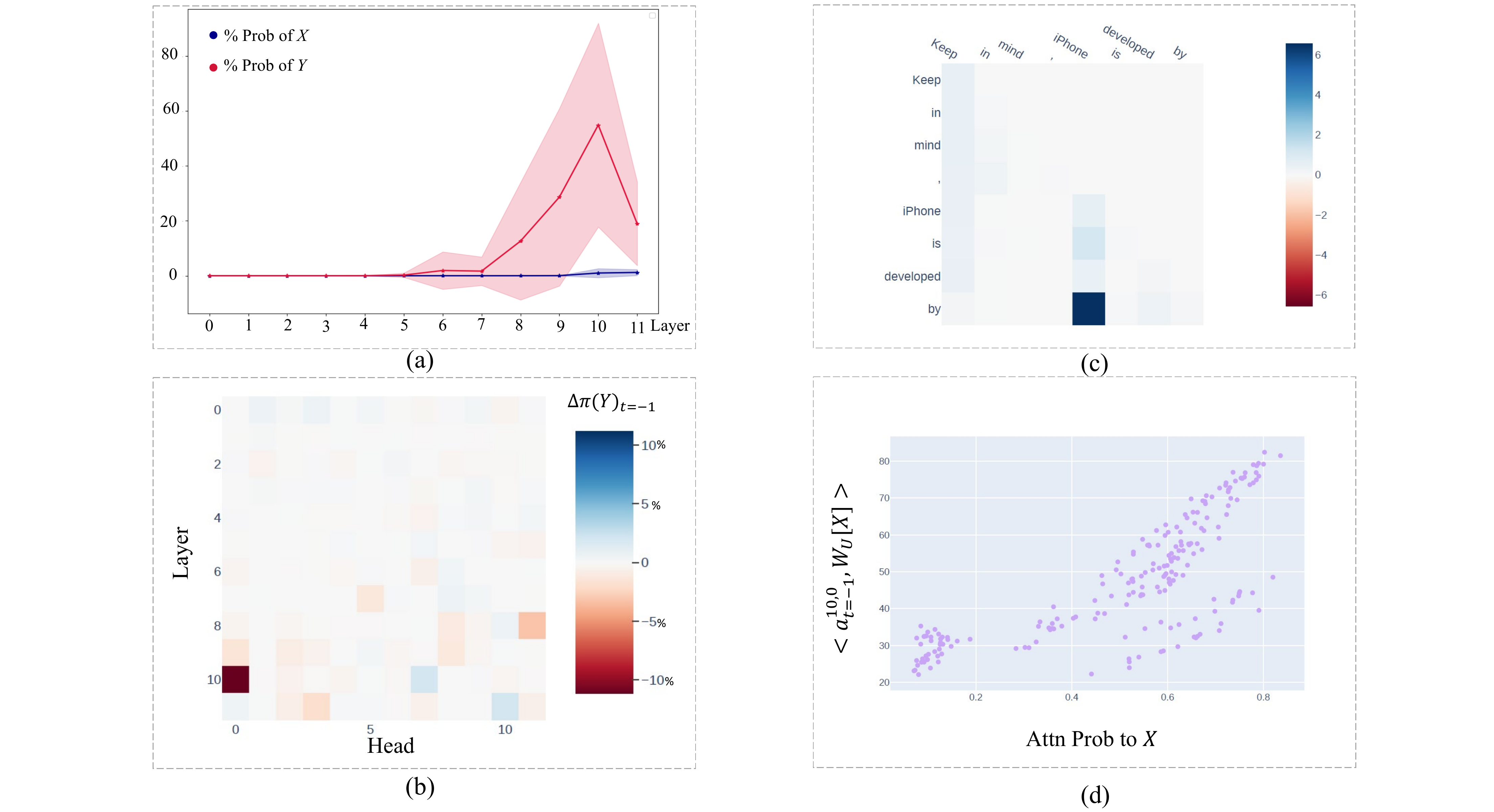}
    \caption{We test GPT-2 small on the product-developer task. (a) The probability of $X$ and $Y$ when early unembed each layer's outputs. (b) The effect of patching path $\textbf{a}^{l,h}_{t=-1}\rightarrow \textbf{r}^{11}_{\text{post}}$. (c) Value weighted attention pattern of L10H0. (d) L10H0 is a mover head.}
    \label{fig:pro-deveffect-to-logits}
\end{figure}

\begin{figure}[t]
    \centering
    \includegraphics[width=\linewidth]{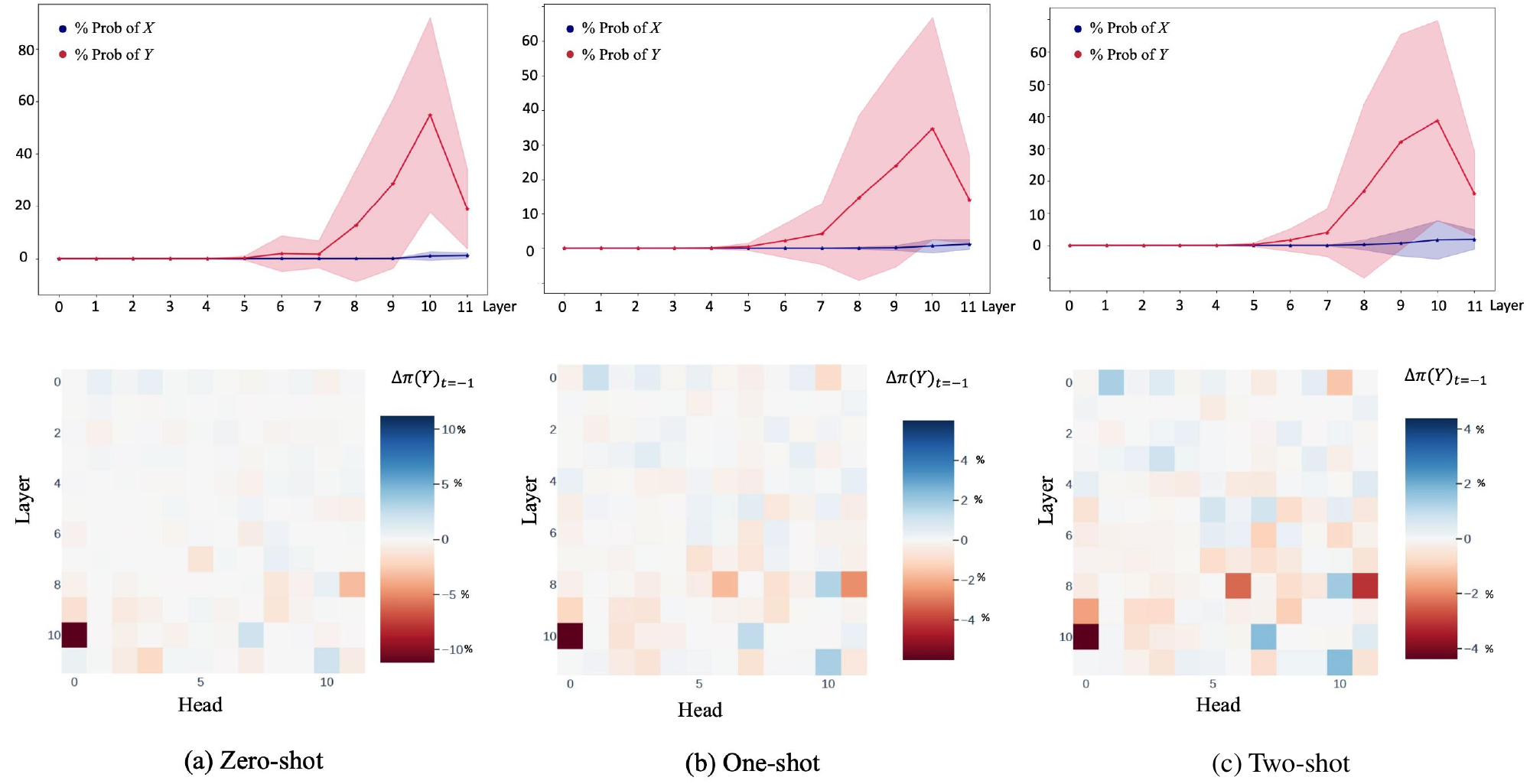}
    \caption{When testing GPT-2 small on the product-developer task, we present probability dynamics of $X$ and $Y$, alongside influential heads impacting final logits across zero (a), one (b), and two-shot (c) settings. 
    The underlying mechanisms utilized by the models remain consistent with what we observed in the country-capital task.}
    \label{fig:pro-devfew-shot}
\end{figure}

\begin{figure}[t]
    \centering
    \includegraphics[width=\linewidth]{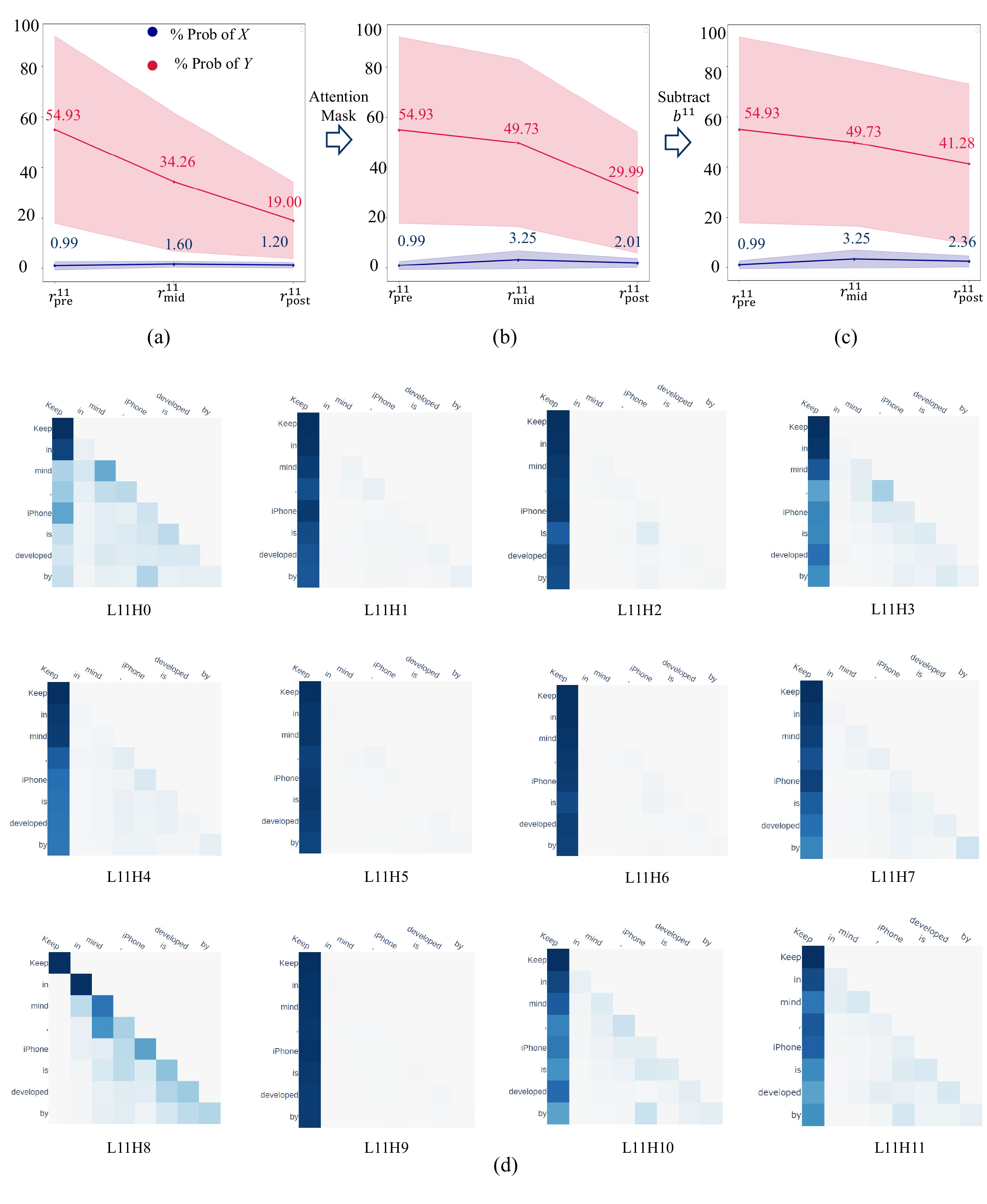}
    \caption{
    Experiments on GPT-2 small in the product-developer task.
    From (a) to (b): The suppression of $Y$ by the attention heads in layer 11 can be eliminated by reallocating their attention. 
    From (b) to (c): The suppression of $Y$ by MLP11 can be mitigated by subtracting $\textbf{b}^{11}$ from the residual stream. 
    The results are obtained in zero-shot settings, and these findings hold true in few-shot settings.
    (d) The attention pattern of heads in layer 11.}
    \label{fig:pro-devlayer11-attn}
\end{figure}

We validated our findings using GPT-2 small on the product-developer task.
We illustrate the probability dynamics of $X$ and $Y$ in Figure \ref{fig:pro-deveffect-to-logits}.
While L10H0 remains activated, L9H8 is not, thus confirming the assertion in the main text that L9H8 is task-specific, whereas L10H0 exhibits a more general activation pattern.
In layer 8, the emergence of probability for $Y$ is evident, indicating the function application.
Our analysis reveals a cosine similarity of 0.635 between $\textbf{b}^{8,\text{proj}}$ and ($\textbf{W}_{U}[Y] - \textbf{W}_{U}[X]$). 
Furthermore, we conducted few-shot experiments on this dataset, depicted in Figure \ref{fig:pro-devfew-shot}. 
Consistent with the observations outlined in the main text, our findings indicate that the mechanisms examined in the few-shot scenarios are still applicable in the zero-shot scenarios.
In Figure \ref{fig:pro-devlayer11-attn}, we observe the anti-overconfidence phenomenon in the final layer. 
This phenomenon can be effectively addressed through our proposed anti-suppression techniques, including applying attention masks and subtracting the intercept from the residual stream.

\clearpage
\section{Experiments on OPT-1.3B}
\label{apx:opt}
\begin{figure}[t]
    \centering
    \includegraphics[width=\textwidth]{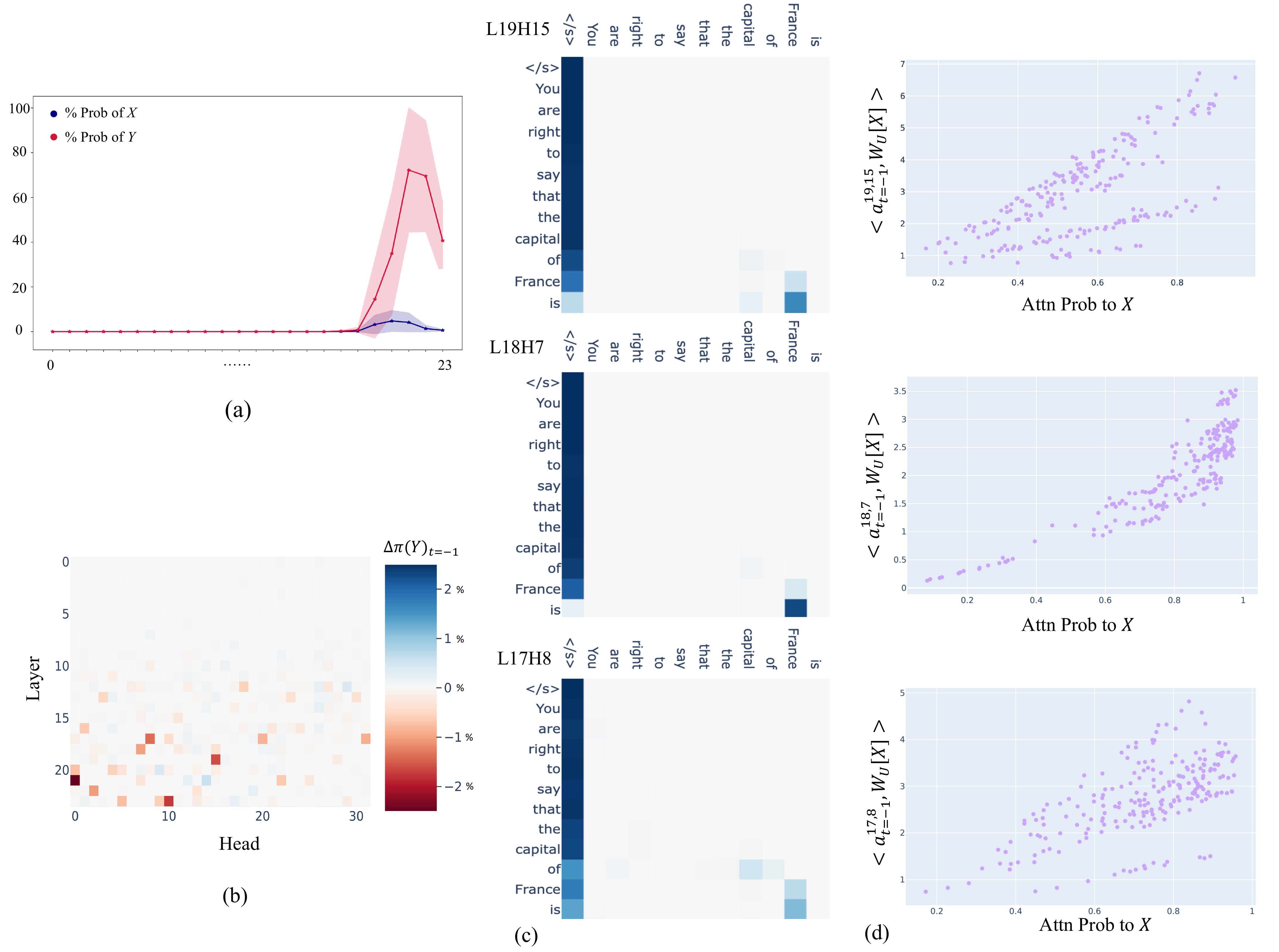}
    \caption{Experiments on OPT-1.3B. 
    The argument passing mechanism and the anti-overconfidence phenomenon are observed.}
    \label{fig:opt-effect-to-logits}
\end{figure}

\begin{figure}[t]
    \centering
    \includegraphics[width=\textwidth]{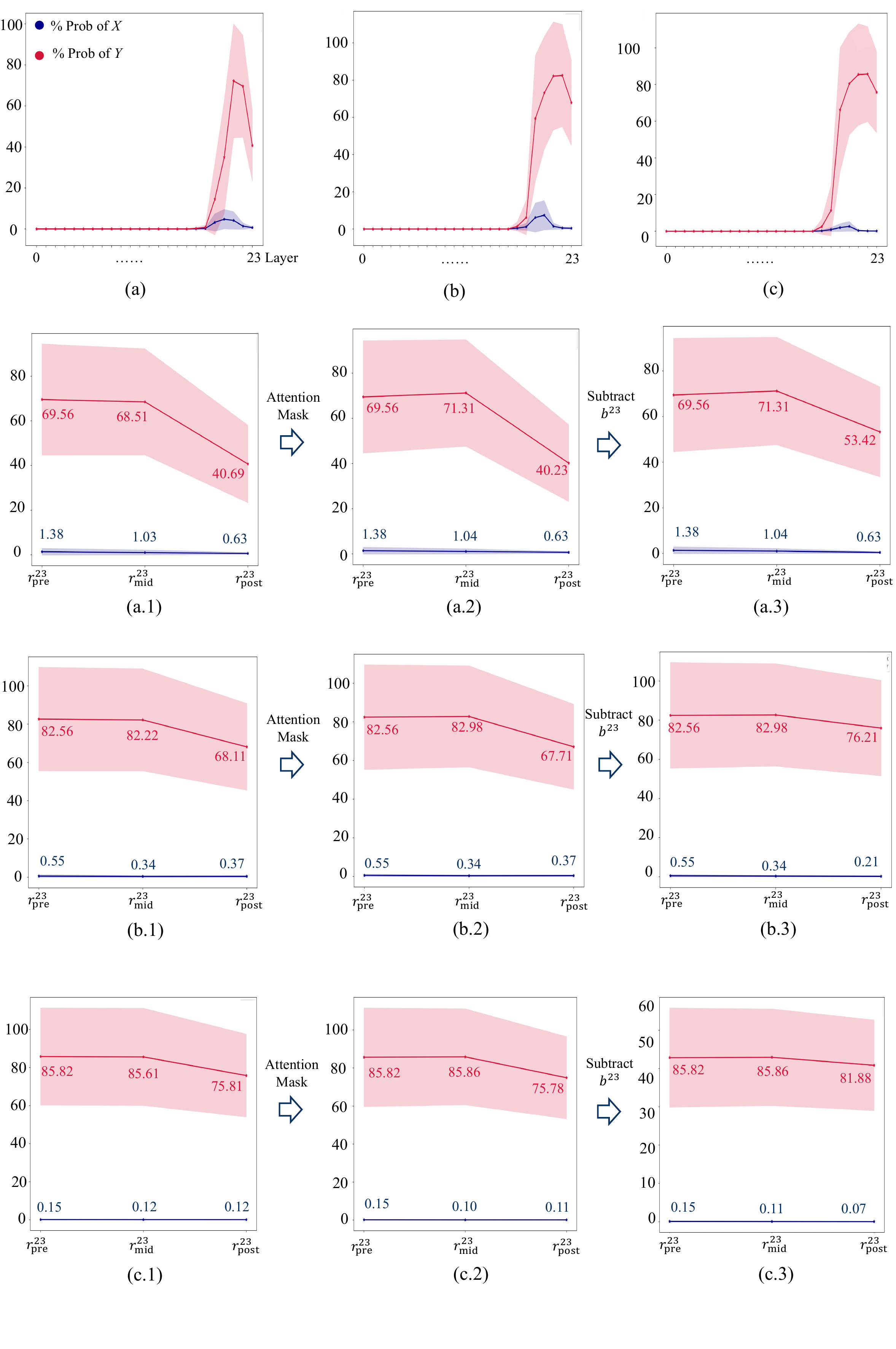}
    \caption{The final layer of OPT-1.3B exhibits suppression of $Y$ across (a) zero-shot, (b) one-shot, and (c) two-shot scenarios.
    Our anti-suppression techniques are applicable for OPT-1.3B, effectively enhancing the probability of $Y$.}
    \label{fig:opt-suppress}
\end{figure}

OPT-1.3B~\cite{zhang2022opt} consists of 24 layers with 32 heads per layer.
In Figure~\ref{fig:opt-effect-to-logits}, we present the probability dynamics across layers, along with path patching experiments.
From Figure~\ref{fig:opt-effect-to-logits}(a), the argument passing is observed in layer 19.
Figure~\ref{fig:opt-effect-to-logits}(b) reveals that L19H15 in the zero-shot settings acts as the first influential argument passer. 
While L17H8 and L18H7 also serve as argument passers, their impact on increasing the probability of $X$ is diminished within the layer.
Their activation is enhanced in few-shots scenarios, making the probability of $Y$ emerge in the residual stream early, as shown in Figure~\ref{fig:opt-suppress}(b) and (c), which is similar to the behavior of L8H11 in GPT-2 small.

Recall that in Section \ref{sec:proj}, two key factors prompted us to realize that $\textbf{b}^{10}$ directs from $X$ to $Y$ within the MLP10 of GPT-2 small:
Firstly, $X$ and $Y$ are equally important tokens that occupy most of the probability distribution.
Secondly, after MLP10, the probability of $X$ decreases, but that of $Y$ increases. 
We have emphasized that these two factors do not always occur, and interpreting the mechanisms of MLPs requires a case-by-case analysis, depending on the model, task, or even prompts.

\begin{table}[h] 
\centering
\renewcommand{\arraystretch}{1.3}
\caption{Probability of several tokens decoded from certain nodes in OPT-1.3B.}
\resizebox{0.4\textwidth}{!}{
    \begin{tabular}{ccc}
    \toprule
        \diagbox{Token}{Node} & $\textbf{r}^{20}_{\text{mid}}$ & $\textbf{r}^{20}_{\text{post}}$ \\\hline
        not & 1.32\% & 4.97\% \\\hline
        London & 4.17\% & 3.21\% \\\hline
        England & 67.40\% & 74.40\%\\\bottomrule
    \end{tabular}}
    \vspace{2mm}
    \label{tab:my_label}
\end{table}
As a case study, given the prompt ``Therefore, it is correct to state that the capital of England is,'' the MLP20 in OPT-1.3B increases the probability of ``London'' by replacing the most likely incorrect answer ``England'' with a safe token ``not,'' rather than directly steering the residual stream towards ``London'' from ``England.''
The cosine similarity between $\textbf{b}^{20,\text{proj}}$ and ($\textbf{W}_{U}[\text{not}]-\textbf{W}_{U}[\text{England}]$) is 0.778, while the cosine similarity between $\textbf{b}^{20,\text{proj}}$ and ($\textbf{W}_{U}[\text{London}]-\textbf{W}_{U}[\text{England}]$) is only 0.1.
As shown in Table~\ref{tab:my_label}, in the transition from $\textbf{r}^{20}_{\text{mid}}$ to $\textbf{r}^{20}_{\text{post}}$, the probabilities of ``London'' and ``not'' are increased by MLP20, while the probability of ``England'' decreases.
This increase in ``London,'' as demonstrated, is merely a by-product of the decline in ``England.''

Regarding the final layer, as illustrated in Figure~\ref{fig:opt-suppress}, OPT-1.3B also demonstrates anti-overconfidence tendencies, suppressing accurate predictions across zero-shot, one-shot, and two-shot scenarios. 
By employing the anti-suppression techniques outlined in Section \ref{sec:few-shot}, we effectively alleviate the anti-overconfidence mechanism employed by OPT.

\section{Experiments on Llama-2-7B and Llama-2-7B-chat}
\label{apx:llama}
\begin{figure}[t]
    \centering
    \includegraphics[width=\textwidth]{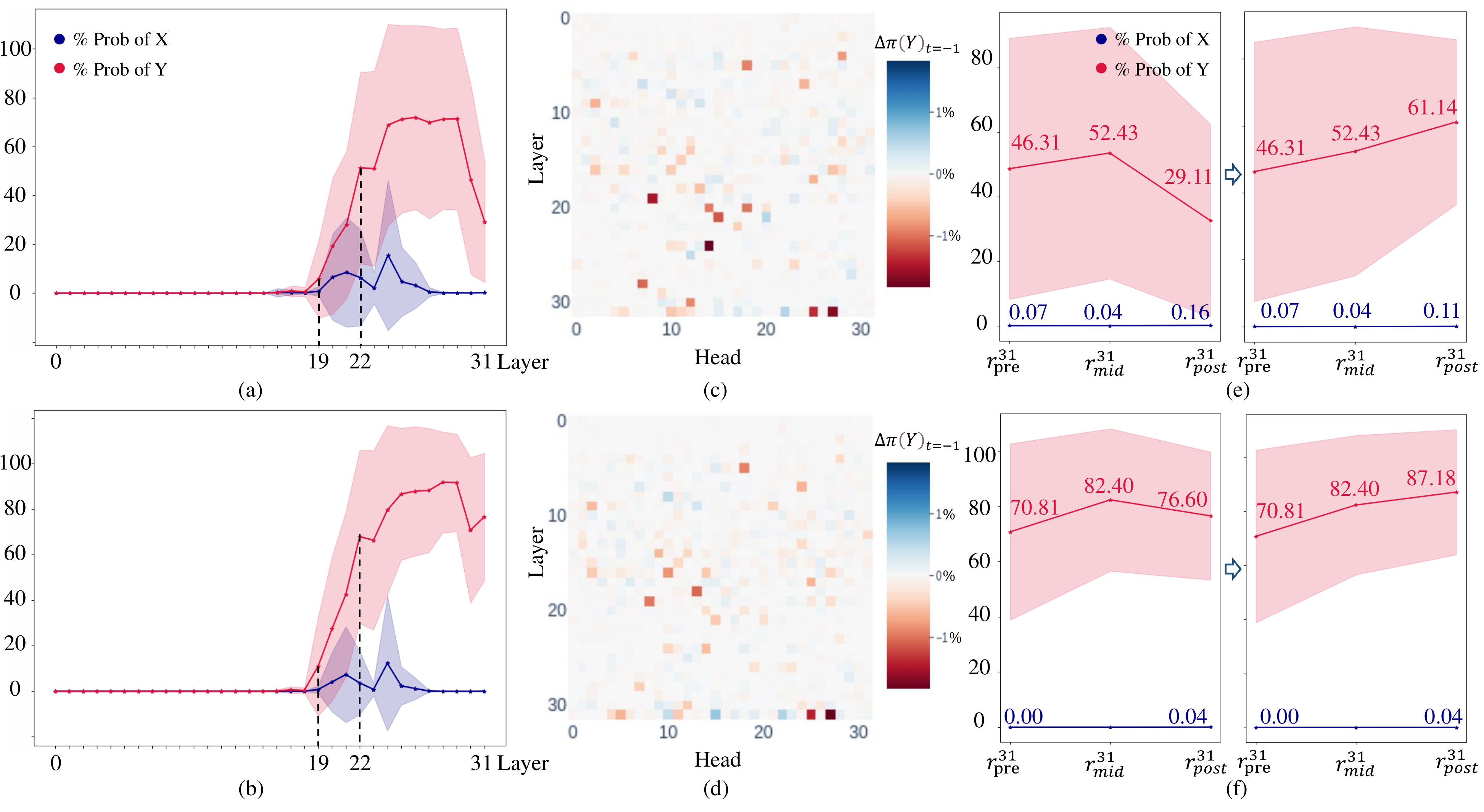}
    \caption{Experimental results on Llama-2-7B and Llama-2-7B-chat. 
    The argument-passing mechanism and the anti-overconfidence phenomenon are observed. Meanwhile, certain behaviors in the chat model differ slightly from those in all other base models studied in this paper.}
    \label{fig:llama-effect-to-logits}
\end{figure}

This section extends our experiments to another popular model family: Llama~\cite{touvron2023llama}, specifically targeting Llama-2-7B and Llama-2-7B-chat. 
Llama-2-7B comprises 32 layers with 32 heads per layer. 
Llama-2-7B-chat is built upon Llama-2-7B, optimized on dialogue data through instruction fine-tuning~\cite{Wei2021FinetunedLM} and RLHF (Reinforcement Learning from Human Feedback~\cite{ziegler2019finetuning}). 
We present the experimental results of both models in Figure~\ref{fig:llama-effect-to-logits}, confirming the validity of the mechanisms above. 
Moreover, we observed behavioral differences between the two models, offering insights into the impact of instruction tuning and RLHF. 
Figure~\ref{fig:llama-effect-to-logits} only presents the zero-shot results, as few-shot experiments yield similar conclusions.

\paragraph{Similar mechanisms shared in two models}
The two models exhibit similar mechanisms for factual recall tasks: 
As shown in Figure~\ref{fig:llama-effect-to-logits}(a) and (b), the ``argument passing'' process occurs at layer 19 in both models.
From Figure~\ref{fig:llama-effect-to-logits}(c) and (d), some mover heads are identified, such as L19H8 and L21H15. 
Two layers are crucial for ``function application'': layer 19 and layer 22. 
In layer 19, the probability of $X$ soars for the first time but the MLP19 makes efforts to output $Y$.
This is reflected in the corresponding cosine similarity values of 0.8822 and 0.8156 between $\textbf{b}^{19,\text{proj}}$ and $\textbf{W}_{U}[Y] - \textbf{W}_{U}[X]$ in two respective models. Meanwhile, in layer 22, where the cosine similarity values are 0.7722 and 0.8637 between $\textbf{b}^{22,\text{proj}}$ and $\textbf{W}_{U}[Y] - \textbf{W}_{U}[X]$, the decline of probability of $X$ and the increase of probability of $Y$ are interpreted.
The MLP in the final layer exhibits the anti-overconfidence mechanism, which can be mitigated by subtracting the respective $\textbf{b}^{31,\text{proj}}$ from the residual stream, as shown in Figure~\ref{fig:llama-effect-to-logits}(e) and (f).
The behavior of the attention heads in the last layer will be discussed below.

\paragraph{Differences: base model versus chat model}
Although both models demonstrate similar factual recall mechanisms, they exhibit divergent behaviors in certain aspects. 
Firstly, Llama-2-7B-chat displays notably higher confidence in achieving a probability of 76.60\% to output $Y$ at the end, in contrast to Llama-2-7B, which yields a probability of only 29.11\%. 
Secondly, attention heads in the final layer of Llama-2-7B (and -chat) positively contribute to accurate predictions, which contrasts with smaller models. 
However, MLPs in the last layer still suppress outputting $Y$ in a manner akin to other models. 

One potential explanation for the first disparity is the impact of instruct tuning and RLHF on language models.
These techniques aim to align the model's output with human preferences, thereby potentially reducing the anti-overconfidence mechanism in chat models' final layer (as it is more sensible for humans to respond given prompts with the capital city' name rather than safe responses). 
Further investigation is necessary to fully comprehend the influence of instruct tuning and RLHF on language model mechanisms, as no prior research has explored this area to our knowledge.
The second disparity stems from the nuanced interaction between active and inactive attention heads in the final layer, with our analysis revealing that only one attention head, L31H8, remains largely inactive, focusing its attention primarily on the initial token—a departure from behaviors observed in smaller models. 
Consequently, the collective effect of attention heads in the final layer exhibits positively.

\section{Details on solving the linear regression}
\label{apx:training}
We present the regression solutions for Eq.\ref{eq:lr} in GPT-2 small on the country-capital task.
Table~\ref{tab:lr-details} presents the learned coefficients for each layer, along with the validation set loss in our 4-fold cross-validation.
The optimization consistently converges to similar solutions for each layer with minimal loss.
Although a few weights in one of the four experiments might diverge from the other three a little, the overall results remain stable enough to support qualitative analysis.
Furthermore, it is natural for the validation MSE loss to rise as the layer depth increases, given that the vectors in deeper layers possess larger norms.
As MSE is a quadratic loss whose penalty is proportional to the square of the error, to alleviate any concerns regarding the convergence, we also present the Absolute Percentage Error (APE) in Table~\ref{tab:lr-relative-error}, expressed as $\frac{\sqrt{\text{MSE}(\Tilde{\textbf{m}}^{l}, \textbf{m}^{l})}}{|| \textbf{m}^{l} ||} \cdot 100\%$.
The APE values for each layer hover around a mere 1\%, indicating minimal deviation.

\begin{table}[ht]
    \centering
    \renewcommand{\arraystretch}{1.3}
    \caption{The learned coefficients for individual layers of GPT-2 small on the country-capital task. 
    Utilizing a 4-fold cross-validation, our results present four rows per layer, accompanied by the corresponding validation loss.}
    \resizebox{\linewidth}{!}{
    \begin{tabular}{ccccccccccccccc}
    \toprule
     \diagbox{Layer $i$}{$w^{l,\cdot}$} &  0 & 1 & 2 & 3 & 4 & 5 & 6 & 7 & 8 & 9 & 10 & 11 & $r$ & val. loss \\\hline
     \multirow{ 4}{*}{0} & -0.3686 & 0.9726 & -0.2864 & 0.2247 & -0.5162 & 1.4244 & -0.3301 & -0.2689 &  0.3259 & -0.0227 & -0.1424 & -0.4358 & -0.0054 & 0.0138\\
     & -0.3800 & 1.0004 & -0.2849 & 0.1797& -0.5336& 1.4090& -0.3203& -0.2905  & 0.3299  & -0.0173 & -0.1103 & -0.4587 & -0.0026 & 0.0126\\
     &-0.3864& 0.9796& -0.2923& 0.2084& -0.5123& 1.4315& -0.3312& -0.2953&  0.2912& -0.0206& -0.1128& -0.4189& -0.0094& 0.0127\\
     &-0.3914& 0.9988& -0.2914& 0.2020& -0.5298& 1.4186& -0.3192& -0.2759&  0.3044& -0.0027& -0.1215& -0.4616& -0.0094& 0.0132\\\hline
    \multirow{ 4}{*}{1} & -0.2393& -0.3280& -0.3935& -0.4470& -0.3156& -0.7519& -0.6252& -0.4797&  -0.4534& -0.7626& -0.3916& -0.4750& -0.1573& 0.0107\\
    &-0.2429& -0.3327& -0.4280& -0.4585& -0.3394& -0.7344& -0.6290& -0.4874&  -0.4745& -0.7808& -0.4168& -0.4807& -0.1531& 0.0117\\
    &-0.2415& -0.3313& -0.3565& -0.4324& -0.3067& -0.7680& -0.6353& -0.4822&  -0.4353& -0.7687& -0.3753& -0.4521& -0.1587& 0.0106\\
    &-0.2372& -0.3275& -0.4528& -0.4507& -0.3318& -0.7301& -0.6256& -0.4961&  -0.4737& -0.7519& -0.4076& -0.4665& -0.1493& 0.0124\\\hline
    \multirow{ 4}{*}{2} & -0.2887& -0.4553& -0.2495& -0.2351& -0.2574& -0.1066& -0.1911& -0.1704&  -0.2661& -0.2948& -0.2853& -0.2810& -0.0932& 0.0153\\
    &-0.3077& -0.4516& -0.2314& -0.2335& -0.2223& -0.0894& -0.2429& -0.1626&  -0.2793& -0.3056& -0.2923& -0.2700& -0.0947& 0.0134\\
    &-0.2853& -0.4341& -0.2446& -0.2390& -0.2337& -0.0997& -0.1758& -0.1613&  -0.2627& -0.2987& -0.2708& -0.2735& -0.0932& 0.0141\\
    &-0.2856& -0.4360& -0.2471& -0.2180& -0.2246& -0.0876& -0.2141& -0.1622&  -0.2542& -0.2926& -0.2864& -0.2580& -0.0961& 0.0147\\\hline
    \multirow{ 4}{*}{3} & -0.8012& -0.1225& -0.1426& 0.0427& -0.6382& -0.3052& -0.0909& -0.1436&  -0.1325& -0.3933& -0.0073& -0.2214& -0.0493& 0.0280\\
    &-0.6559& -0.1240& -0.1452& 0.0240& -0.5122& -0.3043& -0.1142& -0.1450&  -0.1045& -0.3816& -0.0126& -0.2123& -0.0543& 0.0279\\
    &-0.7190& -0.1228& -0.1308& -0.0014& -0.5717& -0.2948& -0.1150& -0.1526&  -0.1014& -0.3853& 0.0165& -0.2102& -0.0517& 0.0267\\
    &-0.6730& -0.1243& -0.1428& 0.0183& -0.5318& -0.3005& -0.1112& -0.1356&  -0.1200& -0.3980& -0.0059& -0.2139& -0.0518& 0.0275\\\hline
    \multirow{ 4}{*}{4} & 0.1505& -0.3024& -0.2439& -0.0427& -0.2223& -0.3436& -0.1989& -0.1019&  -0.4813& -0.2967& -0.1689& -0.0462& -0.0488& 0.0396\\
    &0.1756& -0.3114& -0.2359& -0.0214& -0.2308& -0.3454& -0.1885& -0.1140&  -0.4238& -0.2932& -0.2134& -0.0459& -0.0535& 0.0385\\
    &0.1720& -0.3159& -0.2396& -0.0286& -0.2486& -0.3203& -0.2100& -0.1187&  -0.4806& -0.2973& -0.2572& -0.0410& -0.0516& 0.0408\\
    &0.1771& -0.3127& -0.2385& -0.0200& -0.2402& -0.3337& -0.2011& -0.1045&  -0.4873& -0.2937& -0.2231& -0.0462& -0.0523& 0.0423\\\hline
    \multirow{ 4}{*}{5} &-0.4228& 0.0504& -0.0660& -0.1955& -0.2287& -0.6634& -0.2867& -0.1622&  -0.0024& -0.2363& -0.1871& 0.0556& -0.1288& 0.0560\\
    &-0.5210& -0.0116& -0.0694& -0.1978& -0.2577& -0.7679& -0.2767& -0.1730&  -0.0364& -0.3313& -0.1937& 0.0697& -0.1248& 0.0531\\
    &-0.5281& 0.0226& -0.0544& -0.1991& -0.2338& -0.7687& -0.2494& -0.1651&  -0.0209& -0.3130& -0.1769& 0.0788& -0.1309& 0.0505\\
    &-0.5418& 0.0120& -0.0534& -0.1911& -0.2512& -0.8036& -0.2755& -0.1677&  -0.0075& -0.3269& -0.1885& 0.0743& -0.1289& 0.0476\\\hline
    \multirow{ 4}{*}{6} & -0.3502& -0.3408& -0.1792& -0.2084& -0.2060& -0.2533& -0.1816& -0.2186&  -0.2031& -0.6458& -0.0613& -0.2533& -0.0860& 0.0919\\
    &-0.3403& -0.3274& -0.1782& -0.2105& -0.1974& -0.2533& -0.1714& -0.2152&  -0.2454& -0.6167& -0.1515& -0.2605& -0.0842& 0.1000\\
    &-0.3792& -0.3282& -0.1688& -0.2234& -0.1978& -0.2596& -0.1536& -0.2075&  -0.1982& -0.6181& -0.0502& -0.2456& -0.0829& 0.0931\\
    &-0.3911& -0.3466& -0.1774& -0.2028& -0.2056& -0.2538& -0.1571& -0.2176&  -0.2213& -0.6132& 0.0202& -0.2514& -0.0892& 0.0948\\\hline
    \multirow{ 4}{*}{7} &  0.3063& -0.1772& 0.7709& -0.0521& -0.1195& -0.1144& 0.0212& -0.2405&  0.0831& -0.1142& -0.2198& -1.4439& -0.0679& 0.1322\\
    &0.1489& -0.2222& 0.9735& -0.0497& -0.1247& -0.1020& 0.0019& -0.3483&  0.1054& -0.1258& -0.1997& -1.4224& -0.0637& 0.1356\\
    &0.2591& -0.2214& 0.8430& -0.0717& -0.1161& -0.1049& 0.0296& -0.1952&  0.0712& -0.1193& -0.1429& -1.5894& -0.0678& 0.1313\\
    &0.4062& -0.1981& 0.8108& -0.0660& -0.1272& -0.1004& 0.0159& -0.2716&  0.0753& -0.1178& -0.2205& -1.5409& -0.0726& 0.1333\\\hline
    \multirow{ 4}{*}{8} & 0.0325& -0.6455& -0.0394& -0.0859& -0.0266& -0.0056& -0.3653& 0.1872&  0.1150& -0.1155& -0.1474& -0.1198& -0.0203& 0.2312\\
    &-0.0340& -0.9104& -0.0605& -0.0866& -0.0279& -0.0186& -0.3973& 0.1453&  0.0788& -0.1179& -0.1669& -0.1223& -0.0222& 0.2055\\
    &-0.0393& -0.9561& -0.0617& -0.0975& -0.0263& -0.0183& -0.4420& 0.1786&  0.0936& -0.1270& -0.1670& -0.1288& -0.0213& 0.2093\\
    &-0.0215& -1.0882& -0.0729& -0.0792& -0.0441& -0.0045& -0.4936& 0.1692&  0.0913& -0.1387& -0.1671& -0.1256& -0.0225& 0.2219\\\hline
    \multirow{ 4}{*}{9} & -0.0357& -0.2265& 0.0362& -0.0383& -0.2627& -0.1390& -0.1492& -0.0289&  0.0788& -0.0914& -0.0783& -0.7471& 0.0572& 0.4518\\
    & -0.0456& -0.2074& 0.0455& 0.0299& -0.2481& -0.1098& -0.1293& -0.0514&  0.0820& -0.1197& -0.0763& -0.6468& 0.0548& 0.4441\\
    & -0.0291& -0.2880& 0.0192& -0.0146& -0.3020& -0.1289& -0.1951& -0.0381&  0.0843& -0.0906& -0.0755& -0.6017& 0.0576& 0.4773\\
    & -0.0767& -0.2336& 0.0256& -0.0015& -0.3060& -0.1643& -0.2030& -0.0439&  0.0727& -0.0755& -0.0675& -0.5727& 0.0554& 0.4763\\\hline
    \multirow{ 4}{*}{10} & -0.0693& 0.11717& -0.0888& 0.1557& -0.9873& -0.1603& -0.1341& -0.0967& -0.8532& -0.5571& 0.0060 & -0.3232& 0.0940 & 1.0827 \\
    & -0.0754 & 0.0791 & -0.0852 & 0.1605 & -1.036 & -0.2174 & -0.0937 & -0.1049 & -0.3328 & -0.5843 & 0.0003 & -0.4004 & 0.0978 & 1.1474 \\
    & -0.0780& 0.0764& -0.1289& 0.1524& -0.9804& -0.1736& -0.1194& -0.0984&  -0.8746& -0.5403& 0.0075& -0.4186& 0.1031& 1.1478\\
    & -0.0683 & 0.0347 & -0.0666& 0.1490 & -0.9140 & -0.1260 & -0.1350 & -0.0868 & -1.4332 & -0.45792& 0.0050 & -0.3183 & 0.0954 & 1.3073 \\\hline
    \multirow{ 4}{*}{11} &-0.2254&0.6708&-0.0932&-0.3486&-0.0158&1.7178&0.6173&0.4640& 0.1175&1.1539&0.1363&0.1677&0.0078&1.4323 \\
    &-0.1900&0.8313&-0.0928&-0.3666&0.0761&1.4899&0.5619&0.4020& 0.1213&0.9887&0.1271&0.2024&0.0090&1.5322\\
    &-0.2594&0.8378&-0.0944&-0.3074&-0.0770&1.3838&0.6542&0.4058& 0.1273&0.9250&0.1597&0.2791&0.0071&1.4179\\
    &-0.2315&0.6442&-0.0931&-0.2749&-0.0269&2.0470&0.5061&0.4638& 0.1149&1.1552&0.1235&0.2852&0.0076& 1.4370\\\bottomrule
    \end{tabular}}
    \label{tab:lr-details}
\end{table}

\begin{table}[ht]
    \centering
    \renewcommand{\arraystretch}{1.3}
    \caption{The Absolute Percentage Error (APE) of Eq.\ref{eq:lr} within each layer.}
    \resizebox{\linewidth}{!}{
    \begin{tabular}{ccccccccccccc}\toprule
       Layer &  0 & 1 & 2 & 3 & 4 & 5 & 6 & 7 & 8 & 9 & 10 & 11 \\\hline
       APE & 0.37\% & 1.53\% & 1.71\% & 1.68\% & 1.58\% & 1.63\% & 1.70\% & 1.41\% & 1.49\% & 1.56\% & 0.98\% & 1.65\% \\\bottomrule
    \end{tabular}}
    \label{tab:lr-relative-error}
\end{table}

\begin{figure}[t]
    \centering
    \includegraphics[width=\linewidth]{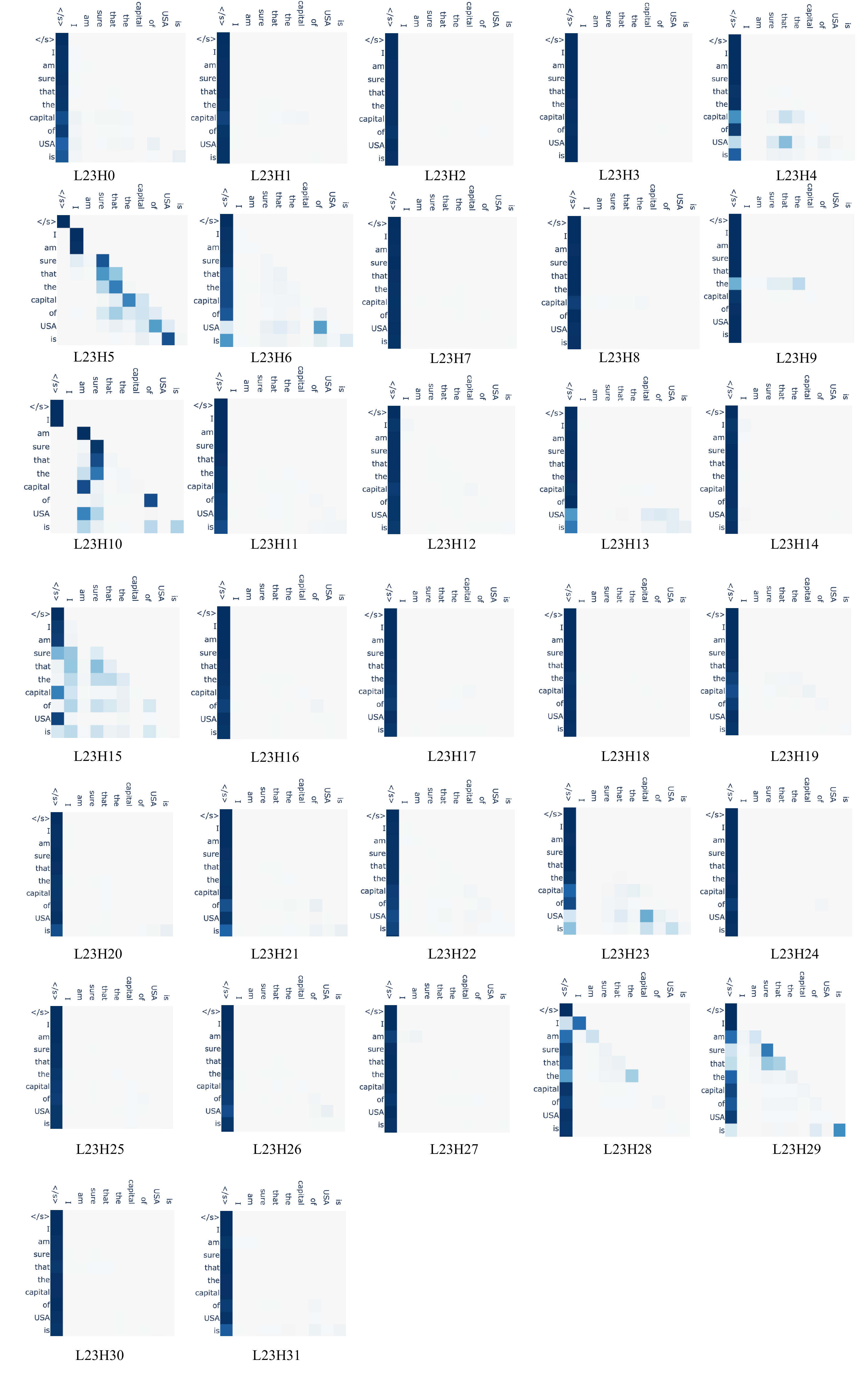}
    \caption{The attention patterns in layer 23 of OPT-1.3B in the country-capital task. }
    \label{fig:opt-23-attn}
\end{figure}

\section{Limitations and future works}
\label{sec:more-questions}
Our paper left several research questions unexplored. 
We present hypotheses for each and leave them for future research endeavours.

Firstly, in Section \ref{sec:attention-pass-arguments}, we directly start with that task-specific heads are activated by task semantics.
However, we omit to explore how the model constructs this task semantics within its shallow layers.
We attempted to trace the information flow from the query and key vectors of the mover heads, only to find that the heads influencing these vectors are scattered, and the individual impact of a head is minimal. 
We posit that the formation of the task semantic occurs through collaborative efforts across multiple circuit paths. 
Thus, attempting to patch a single pathway might be futile, as another pathway could compensate. 
Collaborative circuit paths have long posed a challenge for circuit discovery methods~\cite{conmy2023towards}.
It might be related to the hydra effect in neural networks~\cite {mcgrath2023hydra}.
Additionally, the mechanism of in-context learning boosting model confidence also needs future investigation.

Secondly, the origins of these mechanisms need more research endeavors in the future.
Take, for instance, the behavior of an MLP that redirects the residual stream to its expected answer.
This phenomenon seems to correlate with the residual connections within the Transformer architecture.
When a module handles an input, denoted as $x$, to attain a target, $y$, the most effective strategy entails minimizing loss by structuring the output as $y-x$ instead of aligning the output with the direction of $y$.
This manner is more effective despite the potential for outputs to seem orthogonal or display a negative cosine similarity with $y$.


We recognize that our proposed analysis method, aimed at decomposing MLP outputs, requires preliminary human reasoning. 
Exploring more automated interpretation techniques for MLPs would be a promising research topic.

\section{Broader impacts and safety issues}
This paper reveals many intriguing phenomenons and a novel analysis method to understand MLPs in language models, which have the potential to enhance language models during inference, design new foundational architecture, and improve training paradigms towards a higher level of intelligence.
Moreover, this interpretative endeavour has the potential to steer language models towards alignment with human values, thereby mitigating safety concerns associated with existing large-scale language models. 
Nonetheless, while striving to understand the black-box model fully, there exists a risk that humans may develop models that intentionally violate human values. 
This safety issue needs careful consideration.

\end{document}